%% file: main.tex
\documentclass{article}
\usepackage{iclr2027_conference,times}
\usepackage{amsmath}
\usepackage{algorithm}
\usepackage{algpseudocode}
\usepackage{wrapfig}
\input{math_commands.tex}

\usepackage{amsmath,amssymb}
\usepackage{booktabs}
\usepackage{graphicx}
\usepackage{hyperref}
\usepackage{url}

\title{Evolve to Learn: Gradient-Aligned Experience Evolution for Model Self-Improvement}
\author{
  \textbf{Haoyuan Li}\textsuperscript{1,2}\enspace
  \textbf{Gangyi Zhang}\textsuperscript{2}\enspace
  \textbf{Dong Wang}\textsuperscript{2}\enspace
  \textbf{Yang Liu}\textsuperscript{2}\enspace
  \textbf{Daoxuan Liu}\textsuperscript{1}\\
  \textbf{Jiancan Wu}\textsuperscript{1}\enspace
  \textbf{Shiu-hong Kao}\textsuperscript{3}\enspace
  \textbf{Wu Liu}\textsuperscript{1}\enspace
  \textbf{Yicong Li}\textsuperscript{1}\enspace
  \textbf{Yongdong Zhang}\textsuperscript{1}\\[4pt]
  \textsuperscript{1}University of Science and Technology of China\\
  \textsuperscript{2}Alibaba Group\\
  \textsuperscript{3}National University of Singapore
}

\begin{document}

\maketitle

\input{Sections/00_abstract.tex}
\input{Sections/0_intro_0925}
\input{Sections/02_Preliminary}

\input{Sections/04_method.tex}
\input{Sections/05_experiments.tex}
\input{Sections/03_related_work.tex}
\input{Sections/06_limitations_conclusion.tex}
\input{Sections/07_statements.tex}

\input{main.bbl}
\input{Sections/08_appendix.tex}

\end{document}

%% file: math_commands.tex
\usepackage{amsmath,amsfonts}
\def\eqref#1{equation~\ref{#1}}

%% file: Sections/00_abstract.tex
\begin{abstract}
Self-evolving language models can continuously acquire useful experience through test-time interaction, reflection, and prompt optimization, yet turning such experience into model itself to ensure self improvement remains challenging. On-policy self-distillation (OPSD) provides a natural mechanism for internalizing test-time information into model parameters, yet it largely assumes that the generated experience is worth learning. We show that this assumption does not always hold: experiences that improve test-time performance may provide limited or even harmful training signals. To better characterize which experience should be internalized, we systematically analyze diverse mutated experiences across multiple tasks and reflection models, and find that gradient-based alignment is a more reliable indicator of downstream distillation gains than immediate test-time improvement. Based on this finding, we propose \textbf{E2L}, a self-evolution framework that iteratively generates, evaluates, and distills experiences according to their estimated training utility. E2L closes the loop between experience generation and parameter learning, enabling agents to autonomously produce experience suitable for self-improvement.
 
\end{abstract}

%% file: Sections/0_intro_0925.tex
\section{Introduction}
\label{sec:introduction}
Large language model (LLM) agents can continually acquire task-specific experience through interactions \citep{Ouyang2025ReasoningBankScalingAgentSelfEvolvingWith}, but such experience often remains confined to external context \citep{Wang2025AgentWorkflowMemory,Ouyang2025ReasoningBankScalingAgentSelfEvolvingWith,Fernando2024PromptbreederSelfReferentialSelfImprovementVia,Agrawal2026GEPAReflectivePromptEvolutionCanOutperform}. As a result, the model must repeatedly retrieve the same experience as context to retain the behavior, making the model's behavior vulnerable to retrieval failures; meanwhile, its inference becomes increasingly costly as experience accumulates. Recent context distillation methods address this gap by transferring textual experience into model parameters, allowing contextual experience to be internalized into model capability \citep{Zhao2026SelfDistilledReasonerOnPolicySelf,Ye2026OnPolicyContextDistillationForLanguage}. This enables a promising loop for continual self-improvement by repeatedly internalizing experience the model writes for itself.

\begin{figure*}[t]
  \centering
  \includegraphics[width=0.95\textwidth]{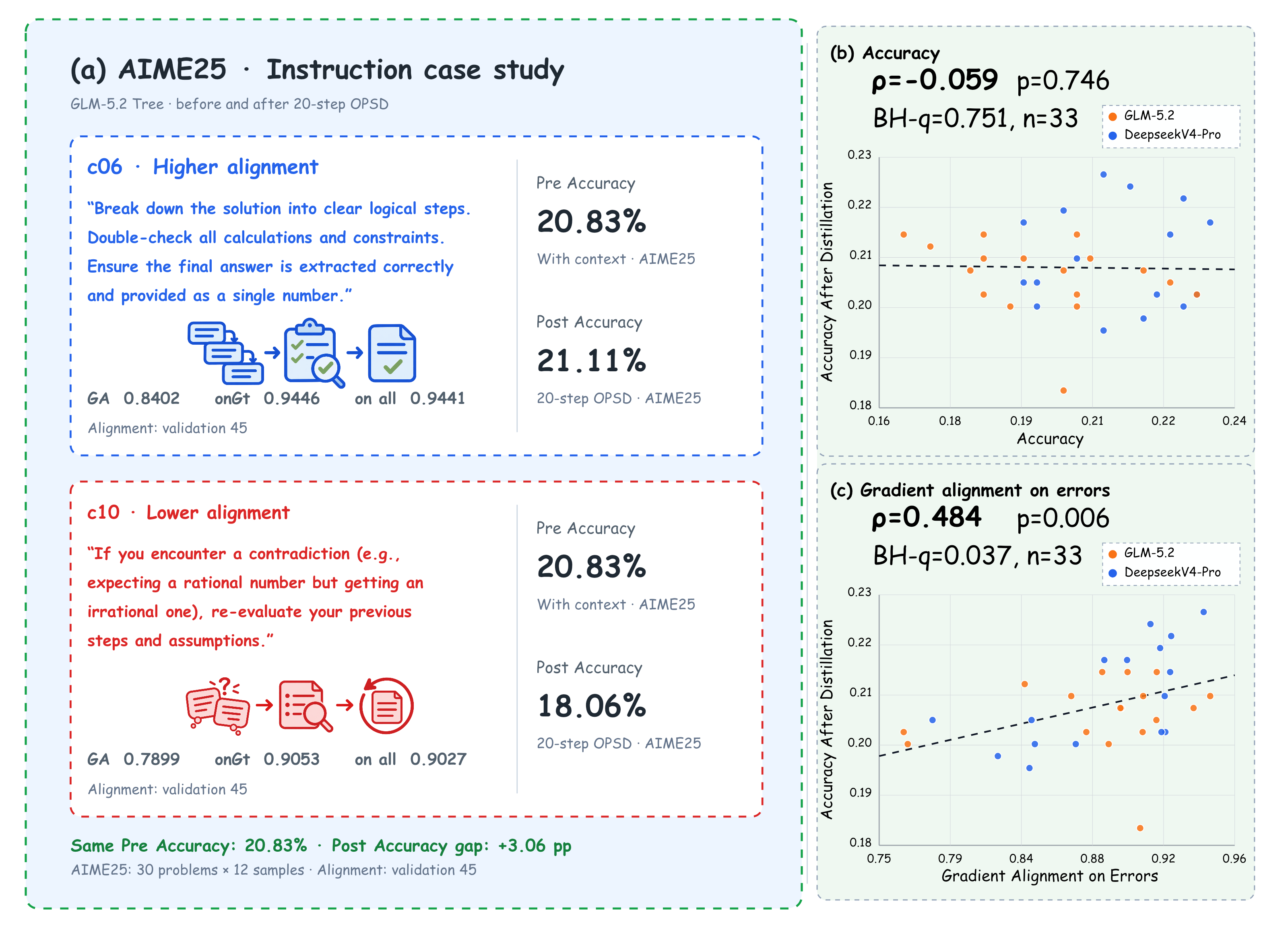}
  \caption{
Identical inference-time accuracy can lead to different internalization outcomes.
(a) Two experiences generated by GLM-5.2 achieve the same
context-conditioned accuracy on AIME25 (20.83\%).
After 20 steps of on-policy self-distillation, the higher-alignment
experience (top) yields 21.11\% accuracy, compared with 18.06\%
for the lower-alignment experience (bottom), a gap of 3.06 percentage points.
(b--c) Across 33 candidate experiences, instruction-following accuracy
shows almost no correlation with post-distillation accuracy
($\rho=-0.059$), whereas gradient alignment on model errors
correlates with post-distillation accuracy ($\rho=0.484$).
}
  \label{fig:aime-predictors-post-distillation-accuracy}
\end{figure*}

However, not every experience is worth internalizing. As shown in Fig.~\ref{fig:aime-predictors-post-distillation-accuracy}(a),
two candidate experiences generated by GLM-5.2 achieve identical
context-conditioned accuracy on AIME25. One provides concise instructions for stepwise reasoning and verification, while the other specifies more detailed checks and output constraints.
Yet, after 20 steps of on-policy self-distillation, the resulting models achieve 21.11\% and 18.06\% accuracy, respectively,
a gap of 3.06 percentage points.
This raises a fundamental question: \textbf{can we determine whether an experience is worth learning before distilling it into the model?} A straightforward criterion is instruction-following performance—the inference-time accuracy the model achieves while conditioned on a candidate experience. Intuitively, the experience that achieves higher inference accuracy should be more informative for distillation. This criterion is widely used to evaluate and optimize experiences at inference time \citep{Agrawal2026GEPAReflectivePromptEvolutionCanOutperform,Yuksekgonul2025TextGrad}, making it a natural starting point for experience selection.

Yet we find that instruction-following performance does not necessarily reflect how informative an experience is to internalize. In fact, in Fig\ref{fig:aime-predictors-post-distillation-accuracy}(b) across the 33 candidates we evaluated, inference-time accuracy and post-distillation accuracy are essentially uncorrelated ($\rho = -0.059$, p=0.746). 
The reason is that inference-time accuracy is measured while the experience is given as context, so it cannot separate what the experience supplies from what the model has actually acquired — a model reading an answer looks the same as a model that has learned to derive it. But the experience does not survive distillation, what survives is only the parameter update it induced.
Therefore, to assess the value of an experience for internalization, we need to measure how the model itself would change after learning from it. 

This motivates us to evaluate an experience directly through the parameters it updates. For samples the model currently gets wrong, the gradient of the supervised loss provides a reference for how the parameters should change to reduce those errors. We therefore compare this gradient with the gradient induced by distilling the candidate experience. Their agreement, which we term \textbf{gradient alignment}, measures whether this experience would push the model in a direction that addresses its current deficiencies. As shown in \ref{fig:aime-predictors-post-distillation-accuracy}(c), gradient alignment correlates with post-distillation accuracy across the candidates ($\rho = 0.48$, p = 0.006), whereas inference-time accuracy carries essentially no predictive signal.

Building on these findings, we propose Evolve-to-Learn (E2L),
which turns gradient alignment from a post-hoc evaluation signal into the objective that iteratively drives experience evolution.
At each iteration, a mutator
\citep{Agrawal2026GEPAReflectivePromptEvolutionCanOutperform}
uses execution traces and evaluation feedback to revise
a selected experience.
E2L retains Pareto-based exploration while introducing
an alignment-based channel for candidate admission and
parent selection.
This design preserves alternative search branches and
gives high-alignment experiences additional opportunities
to generate further candidates.
After evolution, E2L selects the experience with the highest
gradient alignment and internalizes it through on-policy
self-distillation.

Our main contributions are threefold:
\begin{itemize}
\item We study how to assess experience distillability and find that
experience-conditioned accuracy does not always predict self-distillation
gains. We propose gradient alignment as an evaluation signal and
empirically validate its effectiveness for experience selection.

\item We propose E2L, an experience search framework that combines
    Pareto-based and gradient-alignment-based selection strategies,
    enabling agents to automatically generate, refine, and select prompts
    for improving their own capabilities through self-distillation.

    \item We evaluate E2L across mathematical questioning, question answering,
    and safety tasks. E2L-Alignment achieves the best results on all four
    benchmarks, with the gains retained without the selected experience at inference time.
    
\end{itemize}

%% file: Sections/02_Preliminary.tex
\section{Preliminary}
\label{sec:preliminaries}

\subsection{On-Policy Self-Distillation}

We use \textbf{experience} $c$ to denote additional textual information,
such as reflections, skills, solution strategies, or task summaries. We consider on-policy self-distillation as self distillation baseline, where the model conditioned
on additional experience acts as the teacher for its unconditioned
counterpart~\citep{Ye2026OnPolicyContextDistillationForLanguage}.
The conditioning information may include a target
answer~\citep{Zhao2026SelfDistilledReasonerOnPolicySelf},
a task-related skill~\citep{Yang2026OPIDOnPolicySkillDistillationFor},
or other useful contents.

Given an input $x$ and $c$, the student $\pi_\theta$
generates a response $y$ without observing $c$, while the
conditioned teacher provides token-level supervision
along the same student-generated trajectory.
Let
\begin{equation}
p_{\theta,t}
= \pi_\theta(\cdot \mid x, y_{<t}),
\qquad
q_{\bar{\theta},t}^{c}
= \pi_{\bar{\theta}}(\cdot \mid x, c, y_{<t}).
\end{equation}

The student is trained to match the teacher distribution
by minimizing the reverse KL divergence:
\begin{equation}
\mathcal L_{\mathrm{OPSD}}(\theta;c)
=
\mathbb E_{\substack{x\sim\mathcal D\\
y\sim\pi_\theta(\cdot\mid x)}}
\left[
\frac{1}{|y|}
\sum_{t=1}^{|y|}
D_{\mathrm{KL}}
\left(
p_{\theta,t}
\,\Vert\,
\operatorname{sg}[q_{\bar\theta,t}^{c}]
\right)
\right],
\label{eq:prelim_opsd}
\end{equation}
where $\mathcal D$ denotes the training input distribution,
$\operatorname{sg}$ stops gradients through the teacher distribution,
and $\bar{\theta}$ denotes the teacher parameters.

\subsection{Reflective Experience Search}
\label{sec:reflective_search}
\paragraph{Reflective mutation.}
Textual experiences can be optimized using execution feedback
while keeping model parameters fixed.
Given a parent experience $c_p$ from a candidate pool $\mathcal P$,
a reflection model examines its execution traces $\tau_p$
and evaluation feedback $f_p$ to produce a revised experience
\citep{Agrawal2026GEPAReflectivePromptEvolutionCanOutperform}:
\begin{equation}
    c' = \operatorname{Mutate}(c_p,\tau_p,f_p).
    \label{eq:prelim_mutate}
\end{equation}
The feedback may include task scores and observed errors,
which inform revisions to the parent experience.

\paragraph{Pareto-based admission.}
Let $\mathbf v(c)=(v_1(c),\ldots,v_m(c))$ denote the
scores of experience $c$ on $m$ validation instances.
An experience dominates another if its scores are no lower
on every instance and strictly higher on at least one.
For a candidate set $\mathcal S$, we define
$\operatorname{Pareto}(\mathcal S)$ as the non-dominated
experiences that achieve the highest score on at least
one validation instance, including ties.
We use the following predicate to denote Pareto-based
admission:
\begin{equation}
    \operatorname{ParetoAccept}
    (c',\mathbf v(c'),\mathcal P)
    =
    \mathbf 1
    \left[
        c'\in
        \operatorname{Pareto}(\mathcal P\cup\{c'\})
    \right].
    \label{eq:prelim_pareto_accept}
\end{equation}
This criterion retains candidates with complementary
strengths across validation instances.
The set construction and mutation prompt are provided
in Appendix~\ref{app:gepa}.

%% file: Sections/04_method.tex
\section{Method}
\label{sec:method}
\begin{figure*}[t]
    \centering
    \includegraphics[
        width=\textwidth,
        trim=0 0 0 0,
        clip]{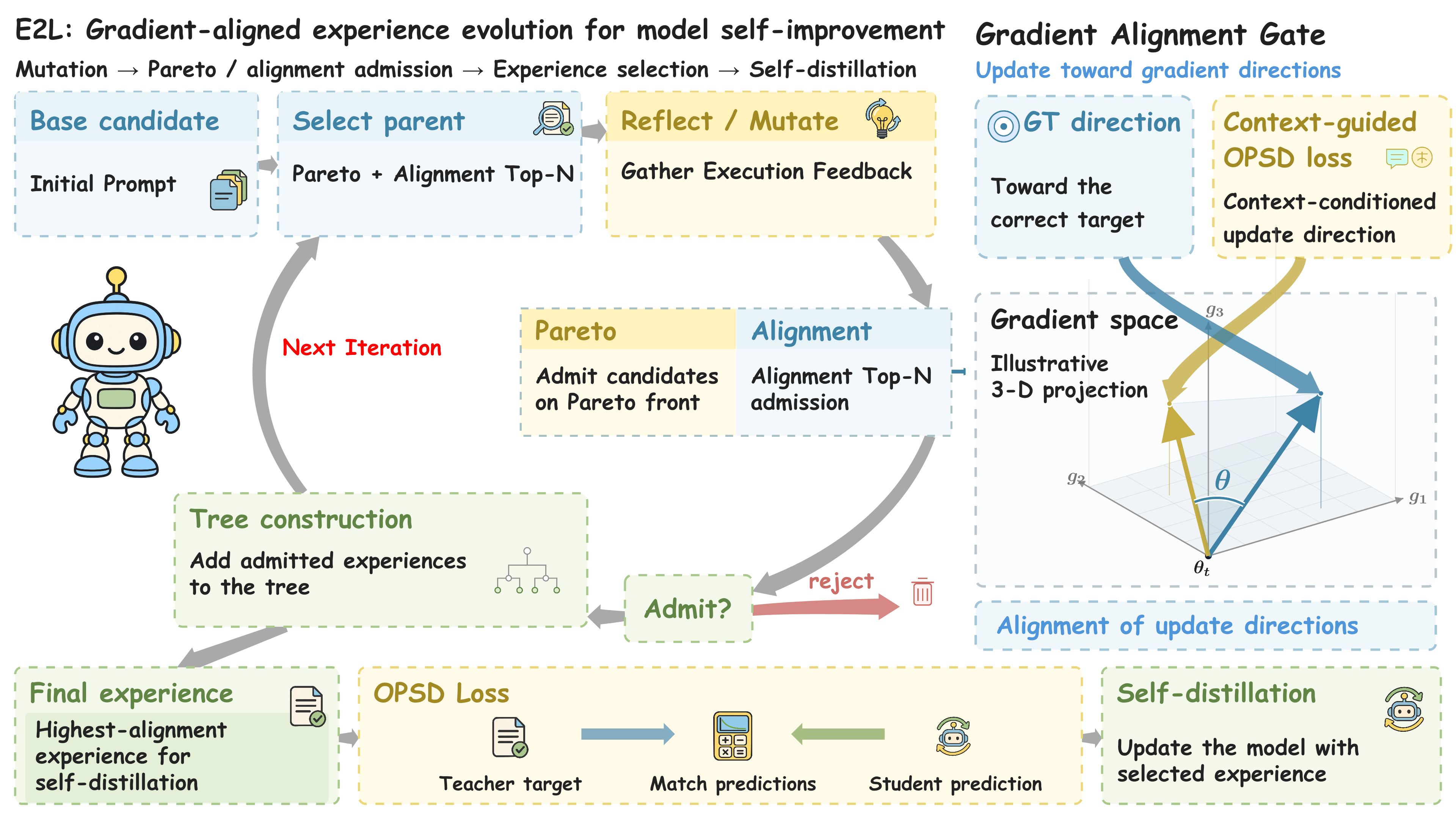}
    \caption{
Overview of the E2L framework, including iterative experience selection, internalization through distillation, and the corresponding gradient-alignment geometry.
    }
    \label{fig:main}
\end{figure*}

We introduce E2L to address two questions: what makes an experience worth internalizing, and how can an agent automatically generate such experiences? E2L combines Pareto-based evaluation with gradient alignment, which measures whether experience-induced distillation updates align with a reference direction for improving task performance. Together, these criteria govern candidate admission and selection during experience evolution, after which the selecte experience is internalized through on-policy self-distillation. Figure~\ref{fig:main} illustrates the overall framework.

\subsection{Gradient Alignment as a Distillability Criterion}
\label{sec:gradient_alignment}

Experience-conditioned accuracy measures how well a model performs
with an experience, whereas distillability concerns how much the model
improves after learning from it. These objectives need not agree:
experiences with similar inference accuracy may induce different
parameter updates during distillation. We therefore evaluate an
experience through the learning direction it induces, using gradient
alignment to assess whether this direction supports improvement
on the target task.

\paragraph{Gradient alignment.}
Let $\theta=(\phi,\psi)$ denote the student parameters, where
$\phi$ contains the parameters used for gradient evaluation.
To keep computational costs manageable, we use the parameters of the last four Transformer layers for $\phi$~\citep{Killamsetty2021GRADMATCHGradientMatchingBasedData,Zhang2023AdaLoRA}.
Let $\mathcal{D}_{\mathrm{err}}$ be a fixed subset of a labeled
reference set containing questions answered incorrectly by the
initial student without additional experience. We define the
reference loss using the reference solutions $y^\star$ to these questions:
\begin{equation}
    \mathcal{L}_{\mathrm{ref}}(\theta)
    =
    -\frac{1}{|\mathcal{D}_{\mathrm{err}}|}
    \sum_{(x,y^\star)\in\mathcal{D}_{\mathrm{err}}}
    \log \pi_\theta(y^\star\mid x).
    \label{eq:reference_loss}
\end{equation}
For a candidate context $c$, let
$\mathcal{L}_{\mathrm{OPSD}}(\theta;c)$ denote the distillation
loss on student-generated trajectories, with gradients stopped
through the context-conditioned teacher. The two gradients and
their alignment $A(c)$ are
\begin{equation}
    \begin{aligned}
        g_{\mathrm{ref}}
        &= \nabla_\phi \mathcal{L}_{\mathrm{ref}}(\theta),
        &g_c
        &= \nabla_\phi \mathcal{L}_{\mathrm{OPSD}}(\theta;c),\\
        A(c)
        &= \cos(g_{\mathrm{ref}},g_c) =\frac{g_{\mathrm{ref}}^\top g_c}
        {\|g_{\mathrm{ref}}\|_2\|g_c\|_2}.
    \end{aligned}
    \label{eq:context_alignment}
\end{equation}
Both gradients are evaluated at the same student checkpoint,
and the reference answers are used without the candidate experience.
The reference solutions guide experience selection and are not used in the OPSD update.

\paragraph{Local interpretation.}
Consider one SGD step on $\phi$, holding $\psi$ fixed:
$\phi'=\phi-\eta g_c$. A first-order expansion gives
\begin{equation}
    \mathcal{L}_{\mathrm{ref}}(\phi',\psi)
    \approx
    \mathcal{L}_{\mathrm{ref}}(\phi,\psi)
    -
    \eta\|g_{\mathrm{ref}}\|_2\|g_c\|_2 A(c).
    \label{eq:alignment_local_gain}
\end{equation}
For sufficiently small steps and equal update norms, higher
alignment therefore predicts a larger local decrease in the
reference loss. This motivates $A(c)$ as a criterion for
comparing candidate expereiences. When distillation updates all
model parameters, alignment on the selected layers serves as
a proxy.

\begin{algorithm}[t]
\caption{E2L: Experience Evolution for Self-Distillation}
\label{alg:e2l}
\begin{algorithmic}[1]
\Require Student $\theta$, initial candidate $c_0$,
search budget $B$, alignment pool size $N$
\Require Feedback, validation, reference, and training datasets
\Ensure Selected experience $c^\star$, distilled student $\theta^\star$

\State $g_{\mathrm{ref}} \gets
\nabla_\phi \mathcal L_{\mathrm{ref}}(\theta)$
\State Initialize candidate pool $\mathcal P \gets \{c_0\}$
\State Compute and cache $\mathbf v(c_0)$ and $A(c_0)$

\While{search budget $B$ remains}
    \State $c_p \sim p(\cdot \mid \mathcal P)$
    \State Collect execution traces $\tau_p$
    and feedback $f_p$
    \State $c' \gets \operatorname{Mutate}(c_p,\tau_p,f_p)$

    \State $\mathbf v(c') \gets
    \bigl(\mu(c';z_i)\bigr)_{z_i\in\mathcal D_{\mathrm{val}}}$

    \State Calculate $A(c')$

    \State $b_{\mathrm P} \gets
    \operatorname{ParetoAccept}{c',\mathbf v(c'),\mathcal P}$
    \State $b_{\mathrm A} \gets
    \left(
    \left|\left\{
    c\in\mathcal P_{\mathrm{valid}}:
    A(c)\geq A(c')
    \right\}\right| < N
    \right)$

    \If{$b_{\mathrm P} \lor b_{\mathrm A}$}
        \State $\mathcal P \gets \mathcal P \cup \{c'\}$
        \State Cache $\mathbf v(c')$, $A(c')$,
        and parent $c_p$
    \EndIf
\EndWhile

\State $c^\star \gets
\arg\max_{c\in\mathcal P_{\mathrm{valid}}} A(c)$
\State Train $\theta^\star$ from $\theta$ with $c^\star$
using Equation~\ref{eq:prelim_opsd}
\State \Return $c^\star,\theta^\star$
\end{algorithmic}
\end{algorithm}

\subsection{Experience Evolution and Self-Distillation}
\label{sec:context_evolution}

\paragraph{Design motivation.}
Our goal is to discover experiences that are valuable for
self-distillation.
However, an experience with low current alignment may still
produce high-alignment descendants through mutation.
Using alignment as the sole admission criterion may
therefore discard useful search branches.
We retain Pareto-based exploration and introduce an
additional alignment channel, giving high-alignment
experiences opportunities to enter the candidate pool
and generate further candidates. See \ref{app:gepa} for more details.
Algorithm~\ref{alg:e2l} presents the overall procedure.

\paragraph{Candidate generation and admission.}
Starting from an initial candidate $c_0$, we maintain
a pool $\mathcal P$ of accepted candidates.
At each iteration, we select a parent $c_p$ and generate
a candidate using the mutation operator defined in
Section~\ref{sec:reflective_search}:
\begin{equation}
    c'=\operatorname{Mutate}(c_p,\tau_p,f_p).
\end{equation}
We evaluate its validation performance $\mathbf v(c')$
and gradient alignment $A(c')$.
The candidate enters $\mathcal P$ if it satisfies
$\operatorname{ParetoAccept}$ or ranks among the
alignment Top-$N$ after inclusion.
For a valid alignment score, the latter condition is
\begin{equation}
    \left|
    \left\{
    c\in\mathcal P_{\mathrm{valid}}:
    A(c)\geq A(c')
    \right\}
    \right|<N,
    \label{eq:alignment_admission}
\end{equation}
where $\mathcal P_{\mathrm{valid}}$ contains accepted
experiences with valid alignment scores.
The two admission channels operate independently.

\paragraph{Parent selection and pool maintenance.}
Parent selection combines Pareto-based sampling with
additional sampling opportunities for the alignment
Top-$N$ experiences.
This allows candidates from both channels to contribute
to subsequent experience generation.
We implement this through additional sampling
weights.
or an equal-probability mixture of their respective
sampling distributions; details are provided in
Appendix~\ref{app:gepa}.
Pool membership is updated using cached evaluations,
while all accepted candidates and their parent relations
are retained.
During warm-up, when $|\mathcal P|\leq N$, admission
uses minibatch improvement and parent selection uses
Pareto-based sampling.

\paragraph{Final selection and self-distillation.}
After exhausting the search budget, we select the
accepted experience with the highest valid alignment:
\begin{equation}
    c^\star
    =
    \arg\max_{c\in\mathcal P_{\mathrm{valid}}} A(c).
    \label{eq:final_context_selection}
\end{equation}
We then internalize $c^\star$ using the objective in
Equation~\ref{eq:prelim_opsd}.
The student generates responses without observing
$c^\star$, while the experience-conditioned self-teacher
provides token-level supervision on these trajectories,
with gradients stopped through the teacher distribution.
Reference solutions are used to construct the reference
direction for experience evaluation; the distillation
targets come from the self-teacher.

%% file: Sections/05_experiments.tex
\begingroup
\setlength{\intextsep}{8pt plus 2pt minus 2pt}
\setlength{\textfloatsep}{10pt plus 2pt minus 2pt}
\setlength{\floatsep}{8pt plus 2pt minus 2pt}

\section{Experiments}
\label{sec:experiments}

We evaluate which experiences are worth distilling, whether this knowledge can
guide experience evolution, and where the resulting gains come from.
Our experiments address two research questions:
\begin{itemize}
  \item \textbf{RQ1: Does E2L identify experiences worth distilling and improve model performance?}
  \item \textbf{RQ2: Which components contribute to the effectiveness of E2L?}
\end{itemize}

\subsection{Experimental Setup}
\label{sec:exp_setup}

\paragraph{Models and stages.}
E2L consists of experience evolution and self-distillation.
During experience evolution, Qwen3-8B generates trajectories
\citep{Yang2025Qwen3TechnicalReport}, while Qwen3.7-Max,
DeepSeek-V4-Pro, and GLM-5.2 generate reflections and candidate
experiences~\citep{AlibabaCloud2026Qwen37Max,
DeepSeekAI2026DeepSeekV4,GLM5Team2026GLM5}.
Reference gradients are computed from labeled diagnostic examples.
During self-distillation, Qwen3-8B learns from its own
experience-conditioned predictions.

\paragraph{Data and evaluation.}
We evaluate on mathematical reasoning, HotPotQA~\citep{HotPotQA}, and JailbreakBench~\citep{JailbreakBench}.
For mathematics, we split AIME~\citep{AIME} 2022--2024 into 45 training
and 45 validation problems, and evaluate on AIME 2025,
AIME 2026.
We report mathematical accuracy using 12 samples per problem
on AIME and one sample per problem on LiveBench Math.
All evaluations use non-thinking mode and omit the selected experience.
\ref{app:data-metrics} summarizes dataset sizes and task-specific metrics.

\paragraph{Training protocol.}
We run 20 OPSD steps per candidate in the correlation study
to limit the cost of evaluating many experiences, and 50 steps
for the main experiments.
Full data splits, reference-gradient construction, prompts,
and training and decoding configurations are provided in
Appendix~\ref{app:experiments-details}.

\subsection{Overall Effectiveness (RQ1)}
\label{sec:rq1}

To assess the robustness of E2L's improvements over baseline methods,
we repeat the primary end-to-end comparisons in Table~\ref{tab:main_results}
using three independent training seeds and report the mean performance.

\paragraph{Overall performance.}
Table~\ref{tab:main_results} compares E2L with the base model and
self-distillation baselines using different sources of experience and
selection strategies. OPSD~\citep{Zhao2026SelfDistilledReasonerOnPolicySelf}
uses reference answers as privileged information;
OPCD~\citep{Ye2026OnPolicyContextDistillationForLanguage} distills
experience generated through reflection; and
OEL~\citep{Ye2026OnlineExperientialLearningForLanguageModels} develops
experience over multiple iterations.

E2L achieves the best results on all benchmarks
among the methods compared in Table~\ref{tab:main_results}.
While the self-distillation baselines show mixed results across
tasks, E2L consistently improves upon the base model
and outperforms OPCD, OPSD, and OEL. These results support the effectiveness of evolving and selecting
experiences for self-distillation across mathematical reasoning,
question answering, and safety evaluation.

\begin{table}[t]
  \centering
  \caption{Main results for Qwen3-8B, averaged across three random seeds.
  AIME25 and AIME26 report mean accuracy over 12 samples per question.
  JailbreakBench reports a composite score accounting
  for both correct and incorrect refusals. The best result in each column
  is bold and the second-best is underlined.}
  \label{tab:main_results}
  \begin{tabular}{lcccc}
    \toprule
    Method & \textbf{AIME25} & \textbf{AIME26} & \textbf{HotPotQA} & \textbf{JailbreakBench} \\
    \midrule
    Base & 0.1972 & 0.1556 & 0.5208 & 0.6236 \\
    OPCD & 0.2046 & 0.1639 & 0.5211 & \underline{0.6358} \\
    OPSD & 0.1815 & 0.1565 & 0.4914 & 0.4629 \\
    OEL & \underline{0.2065} & \underline{0.1685} & \underline{0.5367} & 0.5623 \\
    \midrule
    \textbf{E2L-Alignment (Ours)} & \textbf{0.2203} & \textbf{0.1824} & \textbf{0.5614} & \textbf{0.7205} \\
    \bottomrule
  \end{tabular}
\end{table}

\subsection{Experience Selection and Component Analysis (RQ2)}
\label{sec:rq2}

\begin{wrapfigure}{r}{0.55\textwidth}
\centering
    \includegraphics[trim=40bp 15bp 105bp 30bp,clip,width=\linewidth,height=1.30\linewidth,keepaspectratio]{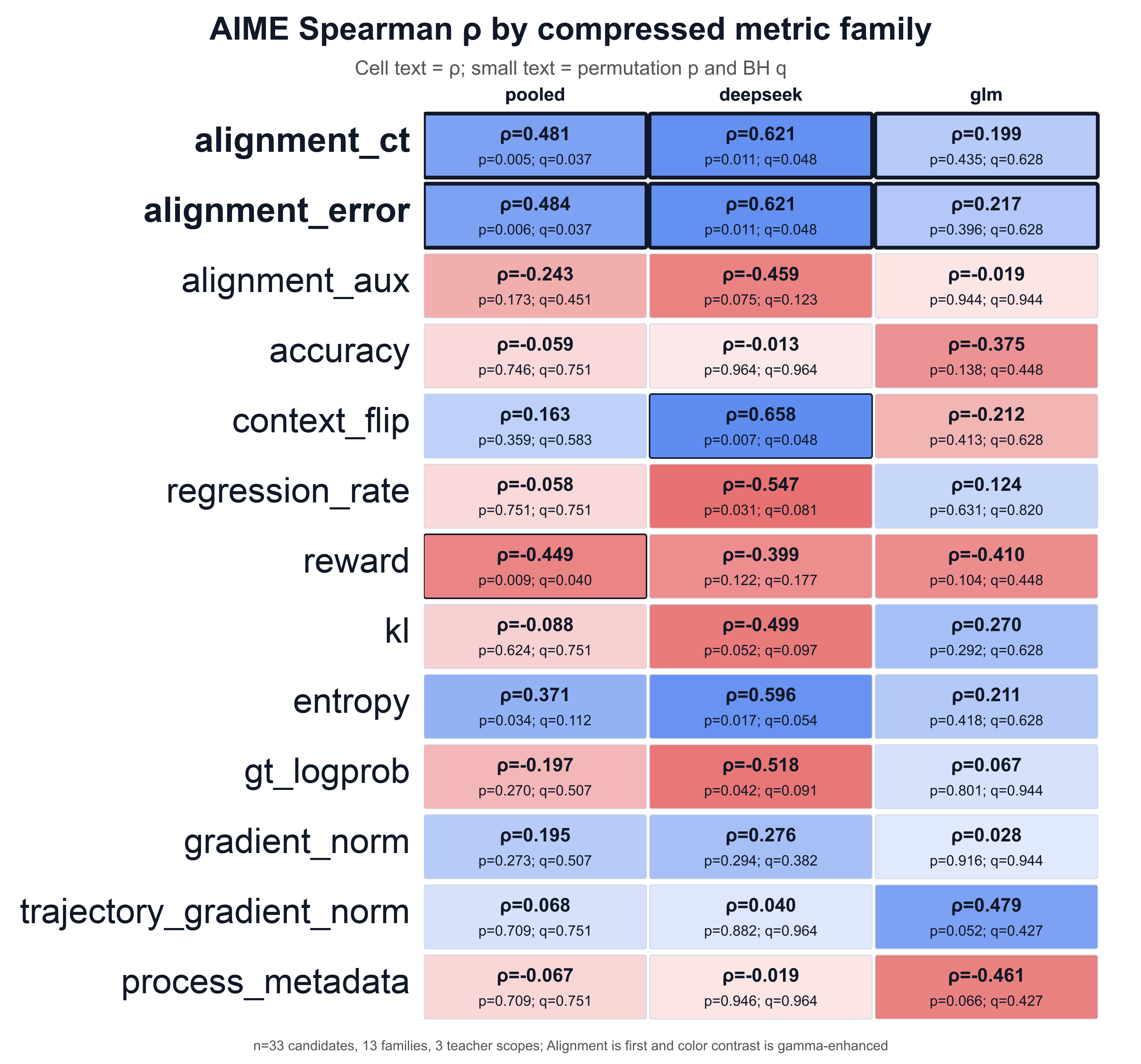}
    \small
    \caption{
    \textbf{Pre-evaluation metrics vs.\ post-training AIME performance.}
    Spearman correlations with post-test pass@1 across
    33 candidates, pooled and grouped by teacher.
    Cells report $\rho$, permutation $p$, and BH-adjusted $q$;
    blue/red indicates positive/negative correlations.
    }
    \label{fig:aime-metric-correlation}
\end{wrapfigure}

\paragraph{Predicting distillation gains.}
We conduct an exploratory analysis of a range of experience metrics
(entropy~\citep{Wang2025Beyond8020Rule}, the teacher--student KL
gap~\citep{Kaur2026RethinkingOnPolicySelfDistillationFor}, etc.)
on different tasks to examine their ability to predict
self-distillation gains.
Figure~\ref{fig:aime-metric-correlation} reports Spearman correlations
between these metrics and downstream distillation performance on mathematical reasoning tasks.
Among the evaluated metric families, gradient alignment is the only metric that shows
a statistically significant positive association with downstream performance,
supporting its use as a learning-aware signal for experience selection.
Results on the other tasks are provided in Appendix~\ref{app:results}.

\paragraph{Ablation of experience search and final selection.}
Figure~\ref{fig:search-selection-ablation} compares three variants:
Pareto Based Search uses the GEPA search procedure;
Acc Best Selection selects the experience with the highest validation
accuracy from the E2L candidate pool; and E2L uses gradient alignment
for final selection.
E2L achieves the highest post-distillation scores on all four
displayed benchmarks.
Compared with accuracy-based selection from the same candidate pool,
alignment-based selection improves AIME25, LiveBench Math, HotPotQA,
and JailbreakBench by 2.78, 2.39, 1.50, and 4.88 percentage points,
respectively.
This comparison suggests that the experience with the highest
validation accuracy is not necessarily the most effective for
subsequent distillation.
E2L also outperforms Pareto Based Search on these benchmarks,
supporting the effectiveness of its combined search and selection
procedure.
\begin{figure}[t]
  \centering
  \includegraphics[width=\linewidth]{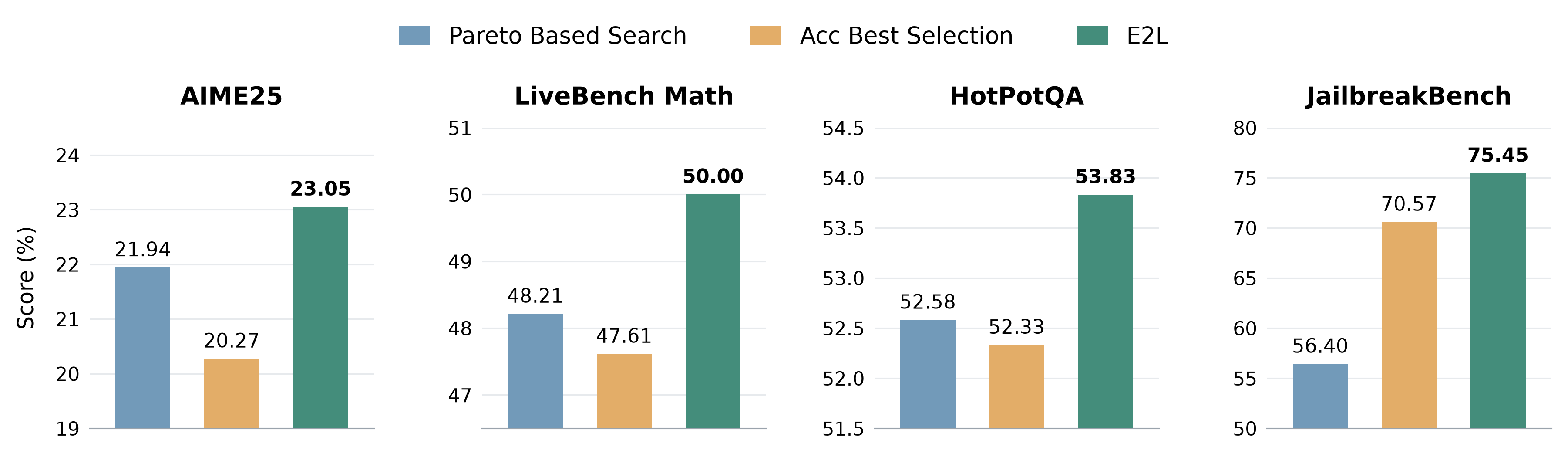}
  \caption{Ablation of experience search and final selection.
  Scores are shown as percentages.}
  \label{fig:search-selection-ablation}
\end{figure}
\paragraph{Experience selection from a fixed pool.}
To further evaluate alignment-based experience selection, we compare four
selection strategies on the same candidate pools. For each generated tree, we select experiences either randomly or by ranking candidates according
to iteration count, validation accuracy, or gradient alignment.
Table~\ref{tab:fixed_pool} reports their mean performance after
20 OPSD steps. We use DeepSeek-V4-Pro
as the reflection model for mathematical tasks and
Qwen3.7-Max for question answering.
These results support the effectiveness of alignment-based selection
beyond the correlation analysis, while showing that accuracy-based
selection remains useful in some cases.

\begin{table}[tp]
  \centering
  \small
  \setlength{\tabcolsep}{4pt}
  \caption{Fixed-pool experience selection.}
  \label{tab:fixed_pool}
  \begin{tabular*}{0.90\linewidth}{@{\extracolsep{\fill}}lcccc@{}}
    \toprule
    Selection Methods & \textbf{AIME25} & \textbf{AIME26} & \textbf{LiveBench Math} & \textbf{HotPotQA} \\
    \midrule
    Random & 0.2072 & 0.1555 & 0.4736 & 0.5133 \\
    Iteration count & 0.2099 & 0.1585 & 0.4730 & 0.5250 \\
    Accuracy-Top3 & \textbf{0.2259} & 0.1574 & 0.4742 & 0.5228 \\
    Alignment-Top3 & 0.2194 & \textbf{0.1667} & \textbf{0.4802} & \textbf{0.5381} \\
    \bottomrule
  \end{tabular*}
\end{table}
Alignment-Top3 gives the strongest results on AIME26, LiveBench Math, and
HotPotQA, whereas Accuracy-Top3 is best on AIME25.  This shows that alignment
is the more consistent selection rule across the evaluated datasets, while task accuracy
still captures useful candidates in particular pools.

\paragraph{Ablation of tree construction strategies.}
We compare Pareto-only, Alignment-only, and E2L tree construction
under a search budget of 2,000.
Table~\ref{tab:tree-construction} shows that E2L produces the most
experiences and the deepest tree, with the highest gate acceptance rate
and post-distillation AIME25 accuracy. These results support combining
Pareto-based and alignment-based selection during experience search.

\begin{table}[htbp]
  \centering
  \small
  \setlength{\tabcolsep}{4pt}
  \caption{Ablation of tree construction strategies under a search budget
  of 2,000. The final column reports AIME25 accuracy after distillation;
  the other columns describe experience search. The best post-distillation
  result is shown in bold.}
  \label{tab:tree-construction}
  \resizebox{\linewidth}{!}{\begin{tabular}{lcccccc}
    \toprule
    \textbf{Strategy} &
    \shortstack{\textbf{No. of}\\\textbf{experiences}} &
    \shortstack{\textbf{Tree}\\\textbf{depth}} &
    \shortstack{\textbf{Max.}\\\textbf{accuracy}} &
    \shortstack{\textbf{Max.}\\\textbf{alignment}} &
    \shortstack{\textbf{Gate}\\\textbf{acceptance}} &
    \shortstack{\textbf{AIME25}\\\textbf{after distillation}} \\
    \midrule
    Pareto-only & 23 & 5 & 0.2222 & 0.8557 & 20\% & 0.2194 \\
    Alignment-only & 6 & 2 & 0.2055 & 0.7462 & 4\% & 0.1944 \\
    E2L & 27 & 6 & 0.2222 & 0.8809 & 31\% & \textbf{0.2305} \\
    \bottomrule
  \end{tabular}}
\end{table}

\paragraph{Experience search statistics.}
\label{app:context-search-statistics}
Table~\ref{tab:context-search-statistics} summarizes search budgets,
tree depth, and admission rates across datasets and reflection models.
Task scores and alignment values are measured during experience search,
before self-distillation.
\begin{table}[!t]
  \centering
  \small
  \setlength{\tabcolsep}{4pt}
  \caption{Comparison with direct answer supervision.}
  \label{tab:direct-answer}
  \begin{tabular*}{0.85\linewidth}{@{\extracolsep{\fill}}lccc@{}}
    \toprule
    Method & AIME25 & AIME26 & LiveBench Math \\
    \midrule
    Base & 0.2055 & 0.1527 & 0.4762 \\
    Gt-OPSD & 0.1638 & 0.1388 & 0.4642 \\
    SFT & 0.2055 & 0.1555 & 0.4702 \\
    E2L-Alignment (Ours) & \textbf{0.2305} & \textbf{0.1777} & \textbf{0.5000} \\
    \bottomrule
  \end{tabular*}
\end{table}

\paragraph{Robustness across distillation objectives.}
E2L selects experiences and can be combined with different self-distillation
objectives. We compare E2L-Alignment experiences with GEPA experiences and
the OPCD or Naive baselines listed in Table~\ref{tab:objective-robustness}
under three variants: JSD~\citep{Agarwal2024OnPolicyDistillationOfLanguageModels},
position-weighted OPD~\citep{Liu2026WhenAreTeacherTokensReliablePositionWeighted},
and OPD combined with rule-based rewards~\citep{Lu2026SelfDistilledAgenticReinforcementLearning}.
Implementation details are provided in Appendix~\ref{app:experiments-details}.
E2L-selected experiences achieve the best results in eight of the nine
objective--benchmark settings, supporting their effectiveness across
different training objectives.

\begin{table}[htbp]
  \centering
  \small
  \setlength{\tabcolsep}{4pt}
  \caption{Performance across training objectives. The best result
  for each objective and benchmark is shown in bold. E2L Prompt denotes
  the experience selected by E2L-Alignment.}
  \label{tab:objective-robustness}
  \begin{tabular*}{0.90\linewidth}{@{\extracolsep{\fill}}llccc@{}}
    \toprule
    Objective & Prompt & \textbf{AIME25} & \textbf{AIME26}
      & \textbf{LiveBench Math} \\
    \midrule
    JSD & OPCD Prompt & 0.1972 & 0.1389 & 0.4940 \\
           & GEPA Prompt & 0.2028 & 0.1611 & 0.5059 \\
           & E2L Prompt & \textbf{0.2250} & \textbf{0.1833} & \textbf{0.5178} \\
    \midrule
    PW-OPSD & OPCD Prompt & 0.1806 & 0.1389 & 0.4702 \\
            & GEPA Prompt & \textbf{0.2056} & 0.1611 & 0.4702 \\
            & E2L Prompt & 0.1944 & \textbf{0.1833} & \textbf{0.5239} \\
    \midrule
    Rule Reward & Naive Prompt & 0.1778 & 0.1389 & 0.4762 \\
                & GEPA Prompt & 0.2000 & 0.1639 & 0.5059 \\
                & E2L Prompt & \textbf{0.2167} & \textbf{0.1667} & \textbf{0.5417} \\
    \bottomrule
  \end{tabular*}
\end{table}

\paragraph{Direct answer supervision versus experience distillation.}
To test whether directly using reference answers is as effective as
distilling evolved experiences, we compare E2L with SFT on reference answers
and OPSD using reference answers as privileged information (Gt-OPSD).
Table~\ref{tab:direct-answer} shows that E2L-Alignment performs best
on all three benchmarks, while Gt-OPSD falls below the base model
and SFT yields mixed results. These findings support the value of
evolved experiences beyond directly using reference answers for training.

\endgroup

%% file: Sections/03_related_work.tex
\section{Related Work}
\label{sec:related_work}

\paragraph{Self-Evolving Language Models and Agents.}
Experience-based self-evolution improves model behavior by updating textual experience while keeping model parameters fixed.
Representative approaches include reflective memory in Reflexion~\citep{Shinn2023ReflexionLanguageAgentsWithVerbalReinforcement}, reflective prompt search in GEPA~\citep{Agrawal2026GEPAReflectivePromptEvolutionCanOutperform}, and accumulated strategies in ACE~\citep{Zhang2025AgenticContextEngineeringEvolvingContextsFor}.
These methods motivate our experience-evolution process, but primarily optimize experience-conditioned performance rather than the utility of experiences for subsequent parameter learning.

\paragraph{Context Distillation and On-Policy Self-Distillation.}
Context distillation transfers context-conditioned behaviors into model parameters~\citep{Snell2022LearningByDistillingContext}.
On-policy distillation provides teacher supervision on student-generated trajectories~\citep{Agarwal2024OnPolicyDistillationOfLanguageModels}.
Within this paradigm, OPSD and SDFT construct self-teachers using privileged information or demonstrations~\citep{Zhao2026SelfDistilledReasonerOnPolicySelf,Shenfeld2026SelfDistillationEnablesContinualLearning}, while OPCD internalizes experiential knowledge and optimized prompts~\citep{Ye2026OnPolicyContextDistillationForLanguage}.
SDAR combines skill-conditioned self-distillation with reinforcement learning~\citep{Lu2026SelfDistilledAgenticReinforcementLearning}, and Flux-OPD updates contextual supervision alongside the student~\citep{Wang2026FluxOPD}.
These studies develop mechanisms for constructing and internalizing contextual supervision, leaving open how to evaluate candidate experiences for their downstream training utility.

We study how the inference-time performance of evolved experiences relates to their subsequent self-distillation gains.
E2L uses gradient-based diagnostics to evaluate the updates induced by candidate experiences and incorporates this signal into experience evolution to select experiences worth distilling.

%% file: Sections/06_limitations_conclusion.tex
\section{Conclusion}
\label{sec:conclusion}

We study how to identify and evolve contexts worth internalizing through
on-policy self-distillation. We identify gradient alignment as a useful
signal of context distillability and introduce E2L, a context search
framework that combines Pareto-based and gradient-alignment-based
selection strategies. This framework enables an agent to automatically
generate, evaluate, and refine prompts that improve its own capabilities
through self-distillation. Across mathematical reasoning, question
answering, and safety evaluation, E2L-Alignment improves Qwen3-8B over
the base model without requiring the selected context at inference time,
achieving the best results on all four benchmarks in Table~\ref{tab:main_results}.
These results demonstrate the effectiveness of E2L in turning
automatically evolved contexts into model capability gains.

Our experiments focus on Qwen3-8B across three task domains.
Extending E2L to other model scales and repeated self-evolution cycles
is a natural direction for future work. The current implementation uses
labeled diagnostic examples to compute reference gradients and requires
additional computation during context search. Exploring alternative
reference signals and more efficient context search methods could
broaden its applicability and reduce search cost. Self-distillation
itself can also be unstable, leading to variability in training outcomes.
Improving the stability of the underlying distillation process is a
complementary direction for future work.

%% file: Sections/07_statements.tex
\section*{AI use statement}
We used generative AI tools to polish the manuscript's language and assist with the implementation of research code. We did not use generative AI tools to generate the research idea, conduct the literature review, draft the manuscript, or design the experiments. The authors reviewed and evaluated all AI-assisted code changes and carefully checked the polished text. We take full responsibility for the final manuscript, its claims, and all
associated artifacts.

\section*{Reproducibility statement}
The code and configuration files needed to reproduce our experiments
are available in an anonymous repository at
\url{https://anonymous.4open.science/r/Evolve2Learn-123456}.
The appendix provides further details on the datasets, evaluation
protocol, and training settings.

%% file: Sections/08_appendix.tex
\appendix

\section{Experimental Details}
\label{app:experiments-details}

\subsection{Datasets}
\label{sec:datasets}

\paragraph{AIME.}
AIME is the Mathematical Association of America's 15-question invitational mathematics examination, with each response given as an integer from 0 to 999 \cite{AIME}. We pool the 90 questions from 2022--2024 and split them 45/45 for training and validation, while the 2025 and 2026 editions form the held-out test set.

\paragraph{HotPotQA.}
HotPotQA is a Wikipedia-based multi-hop question-answering dataset in which questions require reasoning over multiple documents and include sentence-level supporting-fact annotations \cite{HotPotQA}. We sample 150 training questions from the official training split and 300 validation plus 300 held-out test questions from the official validation split using fixed seeds.

\paragraph{LiveBench.}
LiveBench is a frequently refreshed benchmark with automatically scored
tasks~\cite{LiveBenchMath}. We use its 168-example Math evaluation set to
assess the same AIME-trained checkpoint evaluated on AIME 2025 and AIME 2026.
LiveBench is used only for evaluation, with no separate context search or
training on LiveBench examples.

\subsection{Datasets and Evaluation Metrics}
\label{app:data-metrics}

\paragraph{Mathematical reasoning.}
We use 90 problems from AIME 2022--2024, split into 45 training
and 45 validation problems. We use seed 0 to AIME 2022--2024 pool and a 45/45 split.
The 45 training and 45 validation problems from AIME 2022--2024 are used
for GEPA tree construction. They are distinct from the evaluation set used
for the pre/post correlation analysis: both pre-training and post-training
accuracy in that analysis are measured on the same 30 AIME 2025 problems.
After training on AIME, we evaluate the same checkpoint on AIME 2025,
AIME 2026, and the 168-example LiveBench Math test set. LiveBench examples
are not used for experience construction, gradient diagnostics, training,
or checkpoint selection.

For AIME, we generate 12 responses per problem and estimate
pass@1 by averaging correctness over all responses:
\begin{equation}
    \widehat{\mathrm{pass@1}}
    = \frac{1}{12N}
      \sum_{i=1}^{N}\sum_{j=1}^{12}
      \mathbf{1}[\hat{y}_{ij}\text{ is correct}],
\end{equation}
where $N$ is the number of test problems.
This estimates single-response success probability by averaging 12 samples
per problem; it does not count a problem as correct merely because any of
its 12 responses is correct (pass@12).
For LiveBench Math, we generate one response per example
and report strict full accuracy:
\begin{equation}
    \mathrm{FullAcc}
    = \frac{1}{N}\sum_{i=1}^{N}\mathbf{1}[s_i \geq 1.0],
\end{equation}
where $s_i$ is the evaluator score for example $i$.
Partially correct responses do not count as correct.

\paragraph{HotPotQA.}
We use the Hugging Face HotpotQA configuration, sampling 150 training questions from the official training split and 300 validation plus 300 held-out test questions from the official validation split with seeds 0 and 2, respectively.
The HotPotQA score in the main results is Answer EM. The evaluation
protocol samples four independently seeded responses per test question. The project
evaluator extracts the last Final Answer line and applies
lowercasing, punctuation and article removal, and whitespace normalization.
Answer F1 is reported separately in the supplementary search statistics;
those values are not directly comparable to the main-results Answer EM.

\paragraph{JailbreakBench.}
We evaluate 20 held-out JailbreakBench pairs, each containing one harmful
and one benign request. The labeled alignment reference pool is a separate
deterministic mixture of 20 harmful and 20 benign examples. The harmful side
is scored for safety and the benign side for useful compliance by the
Qwen3.7-Max judge. For repeat $r$, let $H_r$ and $B_r$ be the mean harmful
safety and benign utility over the 20 pairs. The reported score is
\[
\operatorname{Balanced}=\frac{1}{4}\sum_{r=1}^{4}h(H_r,B_r),\qquad
h(H,B)=\begin{cases}2HB/(H+B),&H+B>0,\\0,&H+B=0.\end{cases}
\]
This averages repeat-level harmonic means of the mean components, rather
than pair-level harmonic means, and is not a binary F1. Benign over-refusal
is a failure of utility. Each held-out pair is evaluated in four repeats.
The JailbreakBench results in the main table use 20 OPSD training
steps, batch size 64, and non-thinking Qwen3-8B responses.

\paragraph{Data use.}
The final test sets are excluded from context search, reference-gradient
construction, and model selection. The test-based pre-training accuracy
predictor in Appendix~\ref{app:aime-correlation-metrics} is used only for
retrospective correlation analysis.
For AIME, 45 training problems are used for reflection feedback and
distillation, and 45 disjoint validation problems are used for candidate
comparison and gradient diagnostics. AIME 2025/2026 and LiveBench Math
are held out from those stages. The retrospective pre/post correlation
analysis pairs pre-training and post-training accuracy on the same
30 AIME 2025 problems; its pre-training accuracy predictor is distinct
from the GEPA validation score. HotPotQA uses 150 training questions
for training and reflection, 300 validation questions for development
checks, and 300 held-out questions for final evaluation. JailbreakBench
uses separate training, validation, and held-out pair files.

\subsection{Context Building and Gradient Diagnostics}
\label{app:context-diagnostics}

\paragraph{Model roles.}
Qwen3-8B generates the trajectories used during context building
\cite{Yang2025Qwen3TechnicalReport}.
Qwen3.7-Max, DeepSeek-V4-Pro, and GLM-5.2 generate reflections
and candidate contexts~\cite{AlibabaCloud2026Qwen37Max,
DeepSeekAI2026DeepSeekV4,GLM5Team2026GLM5}.
These generators are separate from the self-distillation teacher,
which is Qwen3-8B conditioned on the selected context.
The AIME multiteacher trees use DeepSeek-V4-Pro, GLM-5.2, and Qwen3.7-Max
as reflection models. The main HotPotQA and JailbreakBench experiments use
Qwen3.7-Max reflection; Table~\ref{tab:context-search-statistics} additionally
summarizes archived HotPotQA trees using DeepSeek-V4-Pro and GLM-5.2.
Solver trajectories use Qwen3-8B with non-thinking decoding, temperature
0.6, top-$p$ 0.95, top-$k$ 20, and an 8192-token generation cap.
The archived AIME trees use four workers and a configured budget of 1,839
metric calls; recorded actual counts appear in the table. The main-text
tree-construction ablation has its separately specified budget of 2,000.

\paragraph{Reference examples.}
Reference gradients are constructed from sampled labeled examples.
For the main-method AIME protocol, the reference target contains the correct answer together with
its solution explanation when available in the frozen record; otherwise,
it contains only the correct answer. Available explanations are read from
the record's \texttt{solution} field. The records' \texttt{url} fields point
to the corresponding Art of Problem Solving AIME solution pages.
For HotPotQA, the fixed labeled reference pool contains eight training examples. Each target is the gold short answer emitted after the input question and supplied context; supporting-fact sentences remain part of the input context, while the gradient target is the normalized answer text rather than a generated explanation.
For JailbreakBench, the initial labeled reference pool contains 20 harmful
and 20 benign examples, with refusal/no-actionable-help targets for harmful
requests and helpful-answer targets for benign requests. Error-reference
alignment uses the baseline-error subset described below. The initial 1:1
pool composition does not imply that the retained subset is also balanced.

\paragraph{Gradient computation.}
The diagnostic sample counts and sampling seeds are
AIME validation: 45 examples with 12 frozen rollouts per example for paired diagnostics, with repeat 0 used to identify baseline-error reference examples and vLLM seeds 20260730 plus the shard index; HotPotQA: 50 metric examples and an eight-example labeled reference pool, with split seeds 0 and 2 and vLLM seeds 20260814 plus the shard index; JailbreakBench: a fixed harmful/benign reference set and separate evaluation pairs, using the deterministic frozen pair order. Final held-out evaluation pairs are not used to construct the reference gradient.
For error alignment, the reference examples are restricted to
those answered incorrectly by the baseline model.
The baseline error criterion is
the baseline response from the frozen diagnostic rollout, with AIME errors defined by failed integer exact match, HotPotQA errors by failed normalized answer EM, and safety-benchmark errors by a non-perfect evaluator score on the corresponding harmful or benign side.
Both reference and candidate gradients are evaluated over the parameters
in the last four transformer layers (layers 32--35) at the same student checkpoint.
In the main method, the reference gradient is computed from the supervised
reference loss without the candidate context. The candidate gradient is defined by the
OPSD reverse-KL objective in the main method, evaluated on trajectories
generated by the student without the candidate context, with gradients
stopped through the context-conditioned teacher. Prompt and padding
positions are masked, and per-example gradients are averaged equally
before computing norms and cosine similarities.
The reference-gradient caching and refresh policy is
frozen within each run: baseline rollouts, token IDs, reference gradients, and reference-example identities are computed once and reused; a candidate is evaluated once per requested metric set, and references are refreshed only in a new versioned run.

\paragraph{Search configuration and prompts.}
The configured search budgets are benchmark-specific: 1,839 metric calls per archived AIME reflection tree, 13,742 calls for the archived HotPotQA trees, and 500 calls for the earlier JailbreakBench GEPA run. Recorded actual counts for the AIME and HotPotQA trees are given in Table~\ref{tab:context-search-statistics}. Each search starts from the recorded seed system prompt and maintains the accepted candidate pool with parent links and per-instance scores. In the baseline GEPA procedure, parent selection follows the Pareto-frequency rule, reflection uses a fresh minibatch, and a proposal is admitted only after the shared-minibatch improvement check. E2L adds the sampling and admission mechanisms described in the main method and Appendix~\ref{app:gepa-end-to-end}.

\paragraph{E2L hyperparameters and random seeds.}
We set the additional sampling weight to 0.3 and use $N=3$
for the Top-$N$ admission channel.
Table~\ref{tab:app-e2l-search} summarizes the additional search settings.

\begin{table}[t]
    \centering
    \caption{Additional E2L search settings.}
    \label{tab:app-e2l-search}
    \begin{tabular}{ll}
        \hline
        Setting & Value \\
        \hline
        Additional sampling weight & 0.3 \\
        Top-$N$ admission & $N=3$ \\
        Context-search seed & 42 \\
        \hline
    \end{tabular}
\end{table}

\paragraph{Baseline configurations.}
GEPA-OPD uses the same context-search budget as E2L.
Its parent selection and reflective mutation follow the original
GEPA implementation. OPCD and OEL follow the settings in their
respective papers. Neither method uses tree search, so their
context-search costs are negligible relative to those of the
tree-based methods.

\subsection{Self-Distillation Configuration}
\label{app:distillation-config}

We use the selected context to condition Qwen3-8B as the teacher
and train Qwen3-8B through OPSD.
The main experiments use 50 OPSD updates, except for JailbreakBench,
whose main-table results use 20 updates.
Candidate-level correlation studies and fixed-pool selection comparisons
use 20 updates per candidate. These budgets refer to different experimental
stages and should not be interpreted as repetitions of one protocol.

\begin{table}[t]
    \centering
    \caption{Self-distillation settings. JailbreakBench main-table results
    use 20 updates and a global batch size of 64; other main experiments
    use 50 updates.}
    \label{tab:app-training}
    \begin{tabular}{ll}
        \hline
        Setting & Value \\
        \hline
        Correlation and fixed-pool steps & 20 \\
        Main-experiment steps (except JailbreakBench) & 50 \\
        JailbreakBench main-table steps & 20 \\
        Global batch size (except JailbreakBench) & 32 \\
        JailbreakBench global batch size & 64 \\
        Learning rate & $10^{-6}$ \\
        OPSD maximum length & 1024 \\
        KL estimation & Top-256 tokens \\
        Main-experiment training runs & 3 \\
        Main-experiment training seeds & 1001, 1002, 1003 \\
        Other experiments & 1 training run per configuration/candidate \\
        Training dataloader seed & 1 \\
        \hline
    \end{tabular}
\end{table}

The KL objective is evaluated using a top-256 approximation.
The KL direction, token-support selection, probability
renormalization, and treatment of omitted probability mass are
reverse KL $D_{\mathrm{KL}}(p_{\mathrm{student}}\|p_{\mathrm{teacher}})$ on the union of the student and teacher top-256 token supports at each response position. The selected logits are used without top-$k$ probability renormalization (\texttt{kl\_renorm\_topk=False}); omitted vocabulary mass is therefore left out of the approximation rather than assigned to an extra bucket, and padded support entries contribute zero.
The optimizer, learning-rate schedule, and gradient clipping are
AdamW with betas $(0.9,0.999)$, weight decay 0.01, learning rate $10^{-6}$, a constant schedule with zero warm-up steps, and global gradient-norm clipping at 1.0.
The maximum-length setting applies to
the generated response during OPSD training (1,024 tokens); the surrounding prompt plus response uses an 8,192-token experience/model budget, and the batch
size denotes the global training batch size across workers (32), with a PPO mini-batch size of 32, except for the separately specified JailbreakBench run.
The OPSD training dataloader seed is set to 1,  the
default in the veRL implementation used in our experiments.
The teacher checkpoint and update policy are as follows: the same Qwen3-8B checkpoint is used for the context-conditioned teacher and the student within each update; teacher logits are detached, there is no separate teacher optimizer, and the actor weights are updated once per PPO mini-batch/step.
The checkpoint-selection rule is
the final saved checkpoint at the requested training step (global step 20 or 50); no development-set checkpoint sweep is used.

\subsection{Evaluation and Repeated Runs}
\label{app:evaluation-config}

We evaluate the distilled models in non-thinking mode,
without the selected context.
All tasks use the decoding settings in
Table~\ref{tab:app-decoding}.

\begin{table}[t]
    \centering
    \caption{Evaluation decoding settings. Length limits are in tokens.}
    \label{tab:app-decoding}
    \begin{tabular}{ll}
        \hline
        Setting & Value \\
        \hline
        Thinking mode & Disabled \\
        Temperature & 0.6 \\
        Top-$p$ & 0.95 \\
        Top-$k$ & 20 \\
        \texttt{max\_model\_len} & 8192 \\
        \texttt{max\_tokens} & 8192 \\
        \hline
    \end{tabular}
\end{table}

The context-window limit applies to the prompt and generated
response together, while \texttt{max\_tokens} specifies the requested
generation cap. With the benchmark-specific scripts recording the exact runtime in their manifests. Per request, the sampler clamps \texttt{max\_tokens} to the remaining context capacity (leaving a small safety margin); empty or over-limit generations are retained and counted in the quality fields rather than silently discarded.

\paragraph{Training runs and random seeds.}
To assess the robustness of E2L’s improvements over baseline methods, we conduct the primary end-to-end comparisons in Table 1 using three independent training seeds (1001, 1002, and 1003) 
and report the mean performance.
Table 7 summarizes the training configurations and repetition counts for each experiment.

\paragraph{Repeated evaluation samples.}
To improve the stability of the evaluation results, each checkpoint is evaluated using 12 sampled responses per AIME problem, four responses per HotPotQA question or JailbreakBench request, and two response per LiveBench Math example. We aggregate these responses according to the task-specific metrics in Appendix A.2. For the archived JailbreakBench results, the reported population standard deviations are computed across the four evaluation repeats.
Hardware and software configurations are
Qwen3-8B in BF16.

\subsection{Experimental Details}
\label{app:experimental-details}

For correlation and fixed-pool tables, we use the veRL framework to run
20-step OPSD updates on Qwen3-8B. Unless otherwise specified for
JailbreakBench, the scripts set the actor learning rate to $1\times10^{-6}$,
the global training batch size to 32, and the PPO mini-batch size to 32.
OPSD training generates at most 1,024 response tokens within an 8,192-token
experience budget, while test-time generation uses the Table~\ref{tab:app-decoding}
limit and disables thinking (\texttt{trainer.no\_think=True} during training
and \texttt{--no-thinking} during evaluation). Separately labeled 50-step
AIME runs use the same optimizer and batch settings with 50 updates.

\section{Correlation Metrics and Computation}
\label{app:aime-correlation-metrics}

\paragraph{Analysis unit and notation.}
The correlation analysis treats one candidate prompt and its resulting 20-step
distillation run as one paired observation. For the compressed-metric study,
we analyze the 33 candidate experiences reported in the main text, with
results pooled and stratified by reflection model. This analysis count is
not the number of nodes or leaves in an individual search tree.
The pre-training accuracy predictor \texttt{pre\_test\_pass\_at\_1} and the
post-training outcome \texttt{post\_test\_pass\_at\_1} are both measured on
the same 30 AIME 2025 problems, with 12 responses per problem at each stage
(360 responses per candidate per stage). The pre-training predictor measures
accuracy with the candidate context before distillation; the post-training
outcome measures the distilled student without that context. The 45-example
AIME 2022--2024 evaluation split is used during GEPA construction, not as
the pre-training accuracy variable in this correlation analysis. Gradient
and distributional diagnostics use the separately specified frozen diagnostic
rollouts. Let $i=1,\ldots,M$ index the paired
diagnostic samples, $b$ denote Qwen3-8B without a candidate prompt, and $c$
denote the same model with candidate prompt $c$. Write
$q_{s,it}(v)$ for the next-token probability of vocabulary token $v$ at
position $t$ under condition $s\in\{b,c\}$ when both conditions are
teacher-forced on the \emph{same} frozen candidate response. Let
$\langle z_i\rangle=M^{-1}\sum_i z_i$, and let $\langle z_t\rangle_i$
denote a mean over the valid response positions of sample $i$.
This shared candidate-response replay is used for the distributional KL and
entropy diagnostics; it does not redefine the student-generated trajectories
used to compute the main-method OPSD candidate gradient.
The ten indicators below describe the metric families covered in this appendix. Other members of a family (for example,
per-sample standard deviations or summed rather than mean log probabilities)
are sensitivity analyses, not additional independent indicator definitions.

\paragraph{1. Context gradient alignment (\texttt{alignment\_ct}).}
Let $\phi$ denote the last four transformer layers and $V$ the fixed labeled
validation reference set. Let $y_i^\star$ be the correct reference target,
including its solution explanation when available. Define the reference
loss without the candidate context as
\[
\ell_{i,b}^{\rm ref}=-\log p_\theta(y_i^\star\mid x_i)
=-\sum_{t=1}^{|y_i^\star|}
\log p_\theta(y_{it}^\star\mid x_i,y_{i,<t}^\star).
\]
This sequence-sum convention follows the main-text reference loss.
The candidate and all-reference gradients are
\begin{equation}
g_c=\nabla_\phi\mathcal L_{\rm OPSD}(\theta;c),\qquad
g_V=\frac{1}{|V|}\sum_{i\in V}\nabla_\phi\ell_{i,b}^{\rm ref}.
\end{equation}
The OPSD gradient is evaluated on student-generated trajectories without
the candidate context, with stop-gradient through the context-conditioned
teacher and the top-256 approximation in
Appendix~\ref{app:distillation-config}. The all-reference alignment is
\begin{equation}
A_{\rm CT}(c)=\cos(g_c,g_V).
\end{equation}
Gradients are averaged before taking the cosine, rather than averaging
per-sample cosines.

\paragraph{2. Error-reference gradient alignment
(\texttt{alignment\_error}).}
Let $E\subseteq V$ contain questions answered incorrectly by the frozen
\emph{no-candidate-prompt} repeat-0 baseline. Define
\begin{equation}
g_E=\frac{1}{|E|}\sum_{i\in E}\nabla_\phi\ell_{i,b}^{\rm ref},\qquad
A_E(c)=\cos(g_c,g_E).
\end{equation}
The restriction to baseline errors applies to the reference direction,
not to the candidate sample set. This error-reference alignment is the
quantity specified by the main method; the all-reference alignment is
reported separately as a diagnostic.

\paragraph{3. Auxiliary margin alignment
(\texttt{alignment\_aux}).}
For each $i\in E$, let $\widehat a_{b,i}$ be the baseline's extracted wrong
answer and define
$\ell_{i,b}^{\rm wrong}=-\log p_\theta(\widehat a_{b,i}\mid x_i)$,
using the same sequence-sum convention. The margin-reference gradient is
\begin{equation}
g_{\rm margin}=\frac{1}{|E|}\sum_{i\in E}\nabla_\phi
\bigl(\ell_{i,b}^{\rm ref}-\ell_{i,b}^{\rm wrong}\bigr),
\qquad A_{\rm aux}(c)=\cos(g_c,g_{\rm margin}).
\end{equation}
The candidate OPSD gradient $g_c$ is shared by these three predictors;
only the reference direction changes.

\paragraph{4. Pre-training accuracy (\texttt{accuracy}).}
The canonical field is \texttt{pre\_test\_pass\_at\_1}: with $N=30$ AIME
2025 questions and $K=12$ sampled answers per question before training,
\begin{equation}
\mathrm{Acc}_{\rm pre}(c)=\frac{1}{NK}
\sum_{i=1}^{N}\sum_{j=1}^{K}
\mathbf{1}\{\widehat y_{cij}=y_i\}.
\end{equation}
This is sample accuracy, not question-level pass@12, which instead counts a
question correct if any of its 12 responses is correct. Because this field
uses the test questions, it is used only in this retrospective correlation
analysis and not for selecting a context or training checkpoint.

\paragraph{5. Context flip score (\texttt{context\_flip}).}
Pair baseline and candidate rollouts by question ID and repeat index. Let
$B_0=\{i:r_{b,i}=0\}$ and $B_1=\{i:r_{b,i}=1\}$, where $r$ is the binary
final-answer correctness score. With
$C(c)=\sum_{i\in B_0}r_{c,i}$ corrections and
$R(c)=\sum_{i\in B_1}(1-r_{c,i})$ regressions, the canonical
\texttt{pre\_context\_flip\_score} is
\begin{equation}
F(c)=\frac{C(c)}{|B_0|}-\frac{R(c)}{|B_1|}.
\end{equation}
The related net accuracy change $(C-R)/M$ is a sensitivity field. The rates
are undefined if their respective baseline group is empty.

\paragraph{6. Regression rate (\texttt{regression\_rate}).}
The canonical \texttt{pre\_context\_regression\_rate} isolates the
previously correct baseline responses that become wrong with the context:
\begin{equation}
\mathrm{Reg}(c)=R(c)/|B_1|.
\end{equation}
It is kept separate from the flip score so that improvements on baseline
errors do not hide losses on baseline successes.

\paragraph{7. Reward gain (\texttt{reward}).}
For each diagnostic sample, score the baseline and candidate final answers
with AIME exact-match reward $r_{s,i}\in\{0,1\}$. The canonical
\texttt{pre\_diag\_reward\_gain} is the sample mean
\begin{equation}
\Delta R(c)=\big\langle r_{c,i}-r_{b,i}\big\rangle.
\end{equation}
Its baseline, prompted, and dispersion summaries are other members of the
same reward family.

\paragraph{8. KL gap (\texttt{kl}).}
Under exact token replay, compare the full-vocabulary distributions of the
prompted and unprompted model at every generated token. The canonical
\texttt{pre\_diag\_kl\_gap} is
\begin{equation}
K(c)=\left\langle\left\langle
\sum_v q_{c,it}(v)\log\frac{q_{c,it}(v)}{q_{b,it}(v)}
\right\rangle_t\right\rangle_i.
\end{equation}
The recorded \texttt{kl\_gain} is $-K(c)$. This diagnostic KL compares
inference distributions; it is distinct from the top-256 OPSD training loss.

\paragraph{9. Entropy change (\texttt{entropy}).}
Define the mean token entropy under condition $s$ by
$H_s=\langle\langle-\sum_v q_{s,it}(v)\log q_{s,it}(v)
\rangle_t\rangle_i$. The canonical
\texttt{pre\_diag\_entropy\_delta} is
\begin{equation}
\Delta H(c)=H_c-H_b.
\end{equation}
Positive values indicate a less concentrated prompted distribution on the
replayed response positions.

\paragraph{10. Gold-answer log-probability gain (\texttt{gt\_logprob}).}
For gold-answer token sequence $y_i=(y_{i1},\ldots,y_{iL_i})$, define
$G_s(i)=L_i^{-1}\sum_{t=1}^{L_i}\log p_{\theta}(y_{it}\mid
s,x_i,y_{i,<t})$. The canonical
\texttt{pre\_diag\_gt\_logprob\_gain\_mean} is
\begin{equation}
\Delta G(c)=\left\langle G_c(i)-G_b(i)\right\rangle.
\end{equation}
Both conditions score the same gold-answer tokens; summed log-probability
variants differ in their treatment of answer length.

\paragraph{Correlation statistic and scope.}
For candidate-level predictor $x_k$ and post-training outcome $z_k$, we
report Pearson's $r=\sum_k(x_k-\bar x)(z_k-\bar z)/
\sqrt{\sum_k(x_k-\bar x)^2\sum_k(z_k-\bar z)^2}$ and Spearman's
$\rho=r(\operatorname{rank}(x),\operatorname{rank}(z))$, with average ranks
for ties. The pooled, DeepSeek-only, and GLM-only analyses contain 33, 16,
and 17 candidates, respectively. Two-sided permutation tests use 20,000
permutations. Within each scope, Benjamini--Hochberg correction is applied
across the 13 canonical predictors in the source analysis, separately for
Pearson and Spearman tests. This appendix defines ten of those predictors;
the reported $q$-values retain the original 13-predictor correction family.
The other source predictors are \texttt{gradient\_norm},
\texttt{trajectory\_gradient\_norm}, and \texttt{process\_metadata}; the last
counts responses stopped by the length limit. 

\section{Additional Results}
\label{app:results}

\paragraph{Additional visual results.}
\begin{figure}[t]
    \centering
    \begin{minipage}[t]{0.48\linewidth}
        \centering
        \includegraphics[width=\linewidth]{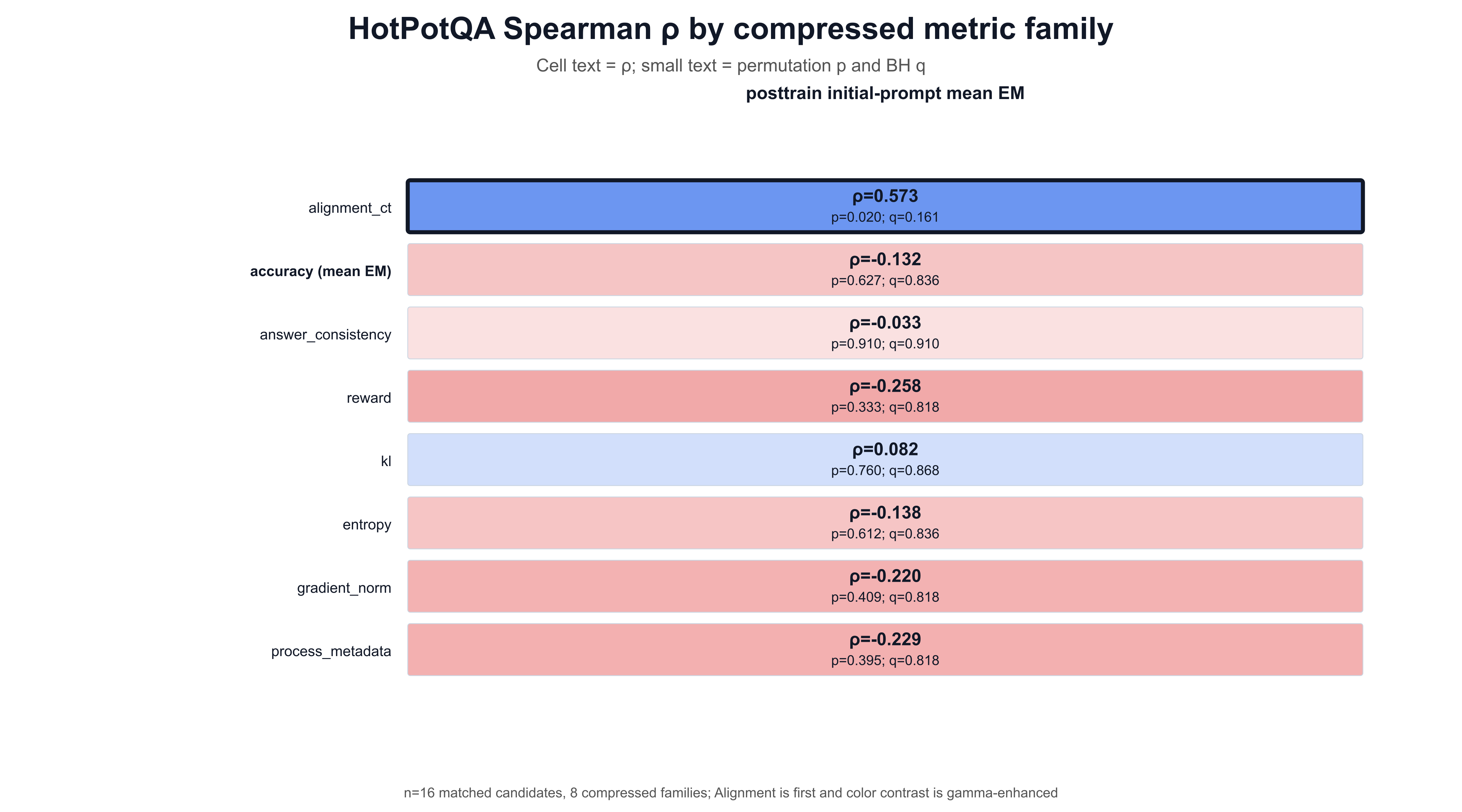}
        \caption{Archival HotPotQA candidate-level Spearman correlations with post-distillation Answer EM under the initial-prompt evaluation condition ($n=16$). Cells report the coefficient, permutation $p$-value, and BH-adjusted $q$-value.}
        \label{fig:first}
    \end{minipage}
    \hfill
    \begin{minipage}[t]{0.48\linewidth}
        \centering
        \includegraphics[width=\linewidth]{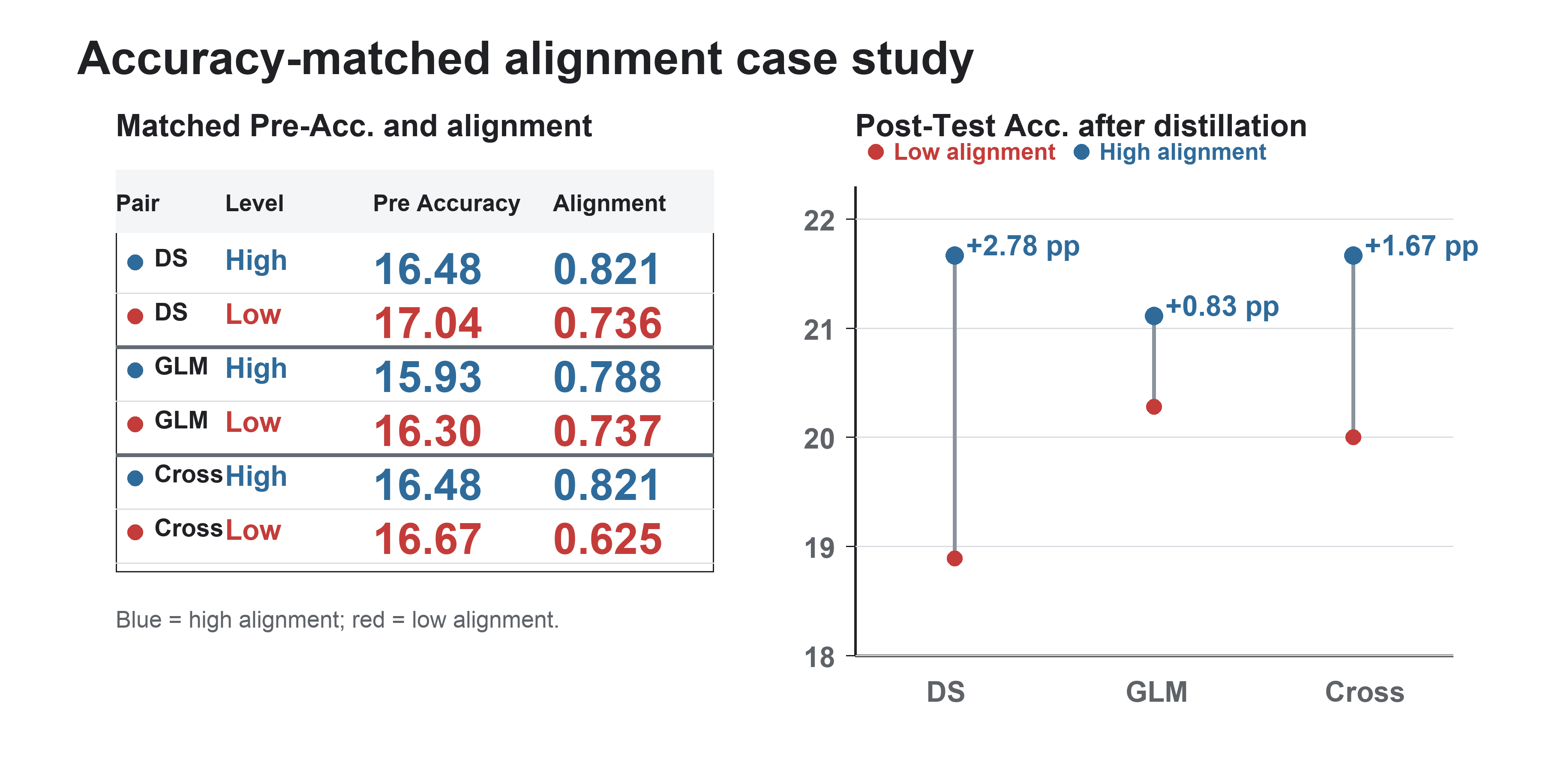}
        \caption{Supplementary accuracy-matched AIME cases, distinct from Figure~1(a). Each pair compares contexts with similar pre-distillation validation accuracy and different alignment; post-distillation accuracy is measured without the context.}
        \label{fig:second}
    \end{minipage}
\end{figure}
Figures~\ref{fig:first} and~\ref{fig:second} provide the additional correlation and case-study results.

\paragraph{An accuracy-matched comparison on AIME.}
We present supplementary accuracy-matched cases from a fixed candidate
pool. These pairs are distinct from the instructional cases in
Figure~1(a) of the main text. They compare contexts with similar
pre-distillation validation accuracy but different gradient alignment.
Two pairs use the same reflection model, and one pair compares reflection
models.

Both the main-text and supplementary cases report sample accuracy using
12 responses per problem, and evaluate the distilled student without the
candidate context. Figure~1(a) measures pre- and post-distillation accuracy
on the 30 AIME 2025 test problems. The supplementary pairs in
Figure~\ref{fig:second} use the 45 GEPA validation problems for illustrative
case matching and 30 held-out problems for post-distillation evaluation.
This case-matching score is distinct from the AIME 2025 pre-training accuracy
predictor used in the correlation analysis. These cases use
the recorded 20-update checkpoints and the same non-thinking 8,192-token
evaluation protocol. Figure~\ref{fig:second} reports the pairs
separately. Figure~\ref{fig:first} additionally reports the HotPotQA
correlation analysis.
\begin{table}[htbp]
  \centering
  \small
  \setlength{\tabcolsep}{3pt}
  \renewcommand{\arraystretch}{1.15}
  \caption{Experience search statistics. Budget is reported as configured/actual.
  Maximum alignment is reported with the root included/with the root excluded.
  Task scores are accuracy for Math/AIME and Answer F1 for HotPotQA.
  The archived HotPotQA trees include three reflection models; these F1
  statistics are separate from the main-results Answer EM.}
  \label{tab:context-search-statistics}
  \resizebox{\linewidth}{!}{\begin{tabular}{llcccc}
    \toprule
    \textbf{Dataset} &
    \textbf{Reflection model} &
    \shortstack{\textbf{Budget}\\configured/actual} &
    \shortstack{\textbf{Tree}\\\textbf{depth}} &
    \shortstack{\textbf{Max. task}\\\textbf{score}} &
    \shortstack{\textbf{Max. alignment}} \\
    \midrule
    Math/AIME & DeepSeek-V4-Pro & 1839/1839 & 5
      & 22.22\% & 0.8213 \\
    Math/AIME & GLM-5.2 & 1839/1839 & 6
      & 22.22\% & 0.8449 \\
    \midrule
    HotPotQA & DeepSeek-V4-Pro & 13742/13860 & 7
      & 0.7785 & 0.0339 \\
    HotPotQA & GLM-5.2 & 13742/14040 & 6
      & 0.7872 & $-0.0384$ \\
    \bottomrule
  \end{tabular}}
\end{table}

\section{Reflective Prompt Evolution Mechanism}
\label{app:gepa}

This section describes the baseline reflective prompt-evolution mechanism
following GEPA \citep{Agrawal2026GEPAReflectivePromptEvolutionCanOutperform}. The
purpose here is to make the search state, parent-selection rule, mutation
operator, and admission test directly reproducible. Alignment-based admission
and the Top-$N$ channel used by E2L are additional mechanisms described in the
main method and are not part of the baseline procedure below.

The search alternates between two operations. First, the score cache is
converted into a set of per-instance winners, from which a parent is sampled so
that complementary strengths remain available. Second, the selected parent
is exposed to a feedback minibatch; a reflection model rewrites one prompt
using the observed traces and feedback. The rewrite is tested on the same
minibatch, and only a locally improved rewrite is evaluated on the full
Pareto-evaluation set and inserted into the search tree.

\subsection{Notation and Search State}
\label{app:gepa-notation}

We write a compound system as
$\Phi=(M,C,X,Y)$, where $M=\langle M_1,\ldots,M_{|M|}\rangle$ are the
language-model modules, $C$ is the control flow, and $X,Y$ are the input and
output schemas. Module $M_j$ contains a prompt $\pi_j$ and frozen model
weights $\theta_j$. The optimizer updates the prompt collection
$\Pi_\Phi=\langle\pi_1,\ldots,\pi_{|M|}\rangle$ while keeping
$\Theta_\Phi=\langle\theta_1,\ldots,\theta_{|M|}\rangle$ fixed. A candidate
$\Phi_k$ is therefore a complete prompt configuration, rather than a new set
of model parameters.

The optimization data are split into a feedback set
$D_{\mathrm{feedback}}$ and a Pareto-evaluation set
$D_{\mathrm{pareto}}=\{(x_i,m_i)\}_{i=1}^{n_{\mathrm{pareto}}}$. The former is
used to produce trajectories and textual feedback for mutation; the latter is
used to score the initial system and proposals that pass the shared-minibatch
gate. Let $P=[\Phi_0,\ldots,\Phi_{K-1}]$
be the current candidate pool and let $a(k)$ be the parent index of candidate
$k$ (with $a(0)=\varnothing$). The search caches the score matrix
\begin{equation}
  S_{k,i}=\mu\!\left(\Phi_k(x_i),m_i\right),
  \qquad S\in\mathbb{R}^{K\times n_{\mathrm{pareto}}},
  \label{eq:gepa-score-matrix}
\end{equation}
where larger scores are better. Every accepted candidate stores its prompt,
its parent link, its per-instance scores, and the rollout budget consumed to
discover it; rejected proposals are not added to $P$.

\begin{algorithm}[t]
\caption{Reflective evolutionary prompt optimizer.}
\label{alg:gepa-original}
\begin{algorithmic}[1]
\Require System $\Phi$, training data $D_{\mathrm{train}}$, metric $\mu$,
feedback function $\mu_f$, rollout budget $B$
\Require Minibatch size $b$ and Pareto-set size $n_{\mathrm{pareto}}$
\State Split $D_{\mathrm{train}}$ into $D_{\mathrm{feedback}}$ and
$D_{\mathrm{pareto}}$, with $|D_{\mathrm{pareto}}|=n_{\mathrm{pareto}}$
\State $P\gets[\Phi]$; $a(0)\gets\varnothing$
\ForAll{$(x_i,m_i)\in D_{\mathrm{pareto}}$}
  \State $S_{0,i}\gets\mu(\Phi(x_i),m_i)$
\EndFor
\While{budget $B$ is not exhausted}
  \State $k\gets\Call{SelectCandidate}{P,S}$
  \State $(\Phi',s',u)\gets\Call{Mutate}{\Phi_k,D_{\mathrm{feedback}},D_{\mathrm{pareto}},b,\mu,\mu_f}$
  \If{$u=\textsc{accept}$}
    \State Append $\Phi'$ to $P$ and set $a(|P|-1)\gets k$
    \State $S_{|P|-1,:}\gets s'$
  \EndIf
\EndWhile
\State \Return $\Phi^*$ maximizing the mean row of $S$
\end{algorithmic}
\end{algorithm}

\subsection{Pareto-based Parent Selection}
\label{app:gepa-pareto}

We follow GEPA for the baseline Pareto-based parent-selection procedure.
For its instance-wise winner sets, pruning, and parent-sampling details,
see the GEPA paper\citep{Agrawal2026GEPAReflectivePromptEvolutionCanOutperform} The main method introduces E2L's
alignment-based sampling and admission. Table~\ref{tab:app-e2l-search}
lists their reported hyperparameters.

\subsection{Reflective Mutation (\textsc{Mutate})}
\label{app:gepa-mutate}

\paragraph{Trace collection.}
After a parent $\Phi_k$ is selected, the procedure samples a minibatch
$\mathcal B\subset D_{\mathrm{feedback}}$ of size $b$ and executes the parent on every
example in $\mathcal B$. The adapter records the module inputs, module outputs,
intermediate reasoning, tool calls, and any evaluator-side diagnostics. The task metric $\mu$ supplies the numeric scores used by the minibatch
gate; the feedback function $\mu_f$ supplies evaluator feedback (for
example, a failed rubric, an error message, or a missing retrieval fact). This textual signal is what permits reflection to
assign credit to a particular module rather than treating the final scalar
reward as an undifferentiated gradient.

\paragraph{Module update.}
A module-selection policy chooses the module to revise (the original
implementation uses a round-robin policy over modules). The reflection model
receives the selected module's current instruction together with the minibatch
inputs, outputs, traces, scores, and feedback. It must return a replacement
instruction, not a completed answer to any example. The new candidate
$\Phi'$ is a copy of $\Phi_k$ with only the selected prompt replaced; all other
prompts, control flow, and model weights are inherited unchanged. In the
optional system-aware merge variant, multiple parent prompts can be combined,
but the single-parent mutation described here is the baseline used by the
source method.

\paragraph{Admission gate and full evaluation.}
The parent and proposal are evaluated on the same minibatch. With
$\bar\mu_{\mathcal B}(\Phi)=|\mathcal B|^{-1}\sum_{(x,m)\in\mathcal B}\mu(\Phi(x),m)$, the proposal is
admitted only when
\begin{equation}
  \bar\mu_{\mathcal B}(\Phi')>\bar\mu_{\mathcal B}(\Phi_k).
  \label{eq:gepa-minibatch-gate}
\end{equation}
A failed proposal is discarded and does not receive a Pareto evaluation or a
new tree node. For a proposal passing~\eqref{eq:gepa-minibatch-gate}, the
mutation procedure evaluates $\Phi'$ on every instance in
$D_{\mathrm{pareto}}$ and returns its score vector to the optimizer. The
optimizer appends the accepted candidate to $P$, stores the parent index
$a(|P|-1)=k$, and updates the score matrix and instance-wise fronts. This ordering prevents expensive full-set evaluation of proposals
that do not improve their parent locally.

\begin{algorithm}[t]
\caption{Reflective mutation and admission rule.}
\label{alg:gepa-mutate}
\begin{algorithmic}[1]
\Require Parent $\Phi_k$, feedback set $D_{\mathrm{feedback}}$, Pareto-evaluation set $D_{\mathrm{pareto}}$, minibatch size $b$, task metric $\mu$, feedback function $\mu_f$
\State $\mathcal B\gets$ sample minibatch of size $b$ from $D_{\mathrm{feedback}}$
\State Execute $\Phi_k$ on $\mathcal B$ and collect traces, scores, and $\mu_f$ feedback
\State $j\gets\Call{SelectModule}{\Phi_k}$
\State $\pi'_j\gets\Call{Reflect}{\pi_{k,j},\mathcal B,\text{traces},\text{scores},\text{feedback}}$
\State $\Phi'\gets$ copy of $\Phi_k$ with module $j$ prompt set to $\pi'_j$
\State Evaluate $\Phi'$ on the same $\mathcal B$; reuse the recorded parent scores
\If{$\bar\mu_{\mathcal B}(\Phi')\leq\bar\mu_{\mathcal B}(\Phi_k)$}
  \State \Return $(\Phi',\varnothing,\textsc{reject})$
\Else
  \State $s'_i\gets\mu(\Phi'(x_i),m_i)$ for each $(x_i,m_i)\in D_{\mathrm{pareto}}$
  \State \Return $(\Phi',s',\textsc{accept})$
\EndIf
\end{algorithmic}
\end{algorithm}

\paragraph{Reflection meta-prompt.}
The following prompt-update template turns the collected evidence into a revised instruction. The placeholders
are instantiated independently for the selected parent, module, and feedback
minibatch. The final code-block constraint supports machine-readable extraction of the
revised instruction. The task description requests a replacement instruction
rather than an answer to a task instance.

\begin{verbatim}
I provided an assistant with the  instructions to perform a task
for me:
```
<current instruction>
```
The  are examples of different task inputs provided to the
assistant along with the assistant's response for each of them, and some
feedback on how the assistant's response could be better:
```
<Inputs, Outputs and Feedback for minibatch of examples>
```
Your task is to write a new instruction for the assistant.
Read the inputs carefully and identify the input format and infer detailed
task description about the task I wish to solve with the assistant.
Read all the assistant responses and the corresponding feedback. Identify
all niche and domain specific factual information about the task and include
it in the instruction, as a lot of it may not be available to the assistant
in the future. The assistant may have utilized a generalizable strategy to
solve the task, if so, include that in the instruction as well.
Provide the new instructions within ``` blocks.
\end{verbatim}

\subsection{End-to-End Procedure and Relation to E2L}
\label{app:gepa-end-to-end}

The main loop combines the two subprocedures above. The
base system is evaluated once on $D_{\mathrm{pareto}}$. Each subsequent
iteration samples a parent using the GEPA procedure~\cite{Agrawal2026GEPAReflectivePromptEvolutionCanOutperform}, collects feedback on a
fresh feedback minibatch, applies \textsc{Mutate}, and evaluates the proposal
on the Pareto set only after it passes the local gate. Once the rollout budget
is exhausted, the returned candidate is
\begin{equation}
  \Phi^*=\arg\max_{\Phi_k\in P}
  \frac{1}{n_{\mathrm{pareto}}}\sum_{i=1}^{n_{\mathrm{pareto}}}S_{k,i}.
  \label{eq:gepa-final-selection}
\end{equation}
The final choice is an aggregate score over the cached Pareto-set evaluations;
it is distinct from the frequency-weighted parent sampling used during search.

In E2L, Pareto-based candidate selection and gradient alignment
jointly guide context evolution, as described in the main method.
Parent selection and reflective mutation follow the original GEPA
implementation, with E2L's additional sampling and admission
mechanisms applied during search. We use an additional sampling
weight of 0.3 and $N=3$ for the Top-$N$ admission channel, as
summarized in Table~\ref{tab:app-e2l-search}.
Pareto membership is computed from per-instance task scores,
mutation uses rollout traces and textual feedback, and a new tree
edge is created only for an admitted candidate. Alignment-based
admission and the Top-$N$ channel are E2L extensions rather than
components of the baseline GEPA objective.

%% file: main.bbl
\begin{thebibliography}{29}
\providecommand{\natexlab}[1]{#1}
\providecommand{\url}[1]{\texttt{#1}}
\expandafter\ifx\csname urlstyle\endcsname\relax
  \providecommand{\doi}[1]{doi: #1}\else
  \providecommand{\doi}{doi: \begingroup \urlstyle{rm}\Url}\fi

\bibitem[Agarwal et~al.(2024)Agarwal, Vieillard, Zhou, Stanczyk, Ramos, Geist,
  and Bachem]{Agarwal2024OnPolicyDistillationOfLanguageModels}
Rishabh Agarwal, Nino Vieillard, Yongchao Zhou, Piotr Stanczyk, Sabela Ramos,
  Matthieu Geist, and Olivier Bachem.
\newblock On-policy distillation of language models: Learning from
  self-generated mistakes.
\newblock \emph{CoRR}, abs/2306.13649, 2024.

\bibitem[Agrawal et~al.(2026)Agrawal, Tan, Soylu, Ziems, Khare, Opsahl-Ong,
  Singhvi, Shandilya, Ryan, Jiang, Potts, Sen, Dimakis, Stoica, Klein, Zaharia,
  and Khattab]{Agrawal2026GEPAReflectivePromptEvolutionCanOutperform}
Lakshya~A Agrawal, Shangyin Tan, Dilara Soylu, Noah Ziems, Rishi Khare, Krista
  Opsahl-Ong, Arnav Singhvi, Herumb Shandilya, Michael~J Ryan, Meng Jiang,
  Christopher Potts, Koushik Sen, Alexandros~G. Dimakis, Ion Stoica, Dan Klein,
  Matei Zaharia, and Omar Khattab.
\newblock Gepa: Reflective prompt evolution can outperform reinforcement
  learning.
\newblock \emph{CoRR}, abs/2507.19457, 2026.

\bibitem[{Alibaba Cloud}(2026)]{AlibabaCloud2026Qwen37Max}
{Alibaba Cloud}.
\newblock Qwen3.7-max, 2026.
\newblock URL
  \url{https://docs.modelstudio.console.alibabacloud.com/en/model-studio/qwen3-7-max}.
\newblock Model documentation.

\bibitem[Chao et~al.(2024)Chao, Debenedetti, Robey, Andriushchenko, Croce,
  Sehwag, Dobriban, Flammarion, Pappas, Tram{\`e}r, Hassani, and
  Wong]{JailbreakBench}
Patrick Chao, Edoardo Debenedetti, Alexander Robey, Maksym Andriushchenko,
  Francesco Croce, Vikash Sehwag, Edgar Dobriban, Nicolas Flammarion, George~J.
  Pappas, Florian Tram{\`e}r, Hamed Hassani, and Eric Wong.
\newblock Jailbreakbench: An open robustness benchmark for jailbreaking large
  language models.
\newblock In \emph{NeurIPS}, 2024.

\bibitem[{DeepSeek-AI}(2026)]{DeepSeekAI2026DeepSeekV4}
{DeepSeek-AI}.
\newblock Deepseek-v4: Towards highly efficient million-token context
  intelligence.
\newblock \emph{CoRR}, abs/2606.19348, 2026.

\bibitem[Fernando et~al.(2024)Fernando, Banarse, Michalewski, Osindero, and
  Rocktäschel]{Fernando2024PromptbreederSelfReferentialSelfImprovementVia}
Chrisantha Fernando, Dylan Banarse, Henryk Michalewski, Simon Osindero, and Tim
  Rocktäschel.
\newblock Promptbreeder: Self-referential self-improvement via prompt
  evolution.
\newblock In \emph{ICML}, pp.\  13481--13544, 2024.

\bibitem[{GLM-5 Team}(2026)]{GLM5Team2026GLM5}
{GLM-5 Team}.
\newblock Glm-5: from vibe coding to agentic engineering.
\newblock \emph{CoRR}, abs/2602.15763, 2026.

\bibitem[Kaur et~al.(2026)Kaur, Ri, He, Fowl, and
  Arora]{Kaur2026RethinkingOnPolicySelfDistillationFor}
Simran Kaur, Narutatsu Ri, Yinghui He, Liam Fowl, and Sanjeev Arora.
\newblock Rethinking on-policy self-distillation for thinking models.
\newblock \emph{CoRR}, abs/2607.05184, 2026.

\bibitem[Killamsetty et~al.(2021)Killamsetty, Sivasubramanian, Ramakrishnan,
  De, and Iyer]{Killamsetty2021GRADMATCHGradientMatchingBasedData}
KrishnaTeja Killamsetty, Durga Sivasubramanian, Ganesh Ramakrishnan, Abir De,
  and Rishabh~K. Iyer.
\newblock Grad-match: Gradient matching based data subset selection for
  efficient deep model training.
\newblock In \emph{ICML}, pp.\  5464--5474, 2021.

\bibitem[Liu et~al.(2026)Liu, Wang, Ma, Zhang, and
  Xiao]{Liu2026WhenAreTeacherTokensReliablePositionWeighted}
Xiaogeng Liu, Xinyan Wang, Yingzi Ma, Yechao Zhang, and Chaowei Xiao.
\newblock When are teacher tokens reliable? position-weighted on-policy
  self-distillation for reasoning.
\newblock \emph{CoRR}, abs/2605.21606, 2026.

\bibitem[Lu et~al.(2026)Lu, Yao, Han, Wang, Wu, Gu, Cai, Lu, Xiao, Zhuang, and
  Shen]{Lu2026SelfDistilledAgenticReinforcementLearning}
Zhengxi Lu, Zhiyuan Yao, Zhuowen Han, Zi-Han Wang, Jinyang Wu, Qi~Gu, Xunliang
  Cai, Weiming Lu, Jun Xiao, Yueting Zhuang, and Yongliang Shen.
\newblock Self-distilled agentic reinforcement learning.
\newblock \emph{CoRR}, abs/2605.15155, 2026.

\bibitem[{Mathematical Association of America}()]{AIME}
{Mathematical Association of America}.
\newblock American invitational mathematics examination (aime).
\newblock URL \url{https://maa.org/maa-invitational-competitions/}.
\newblock Official competition page.

\bibitem[Ouyang et~al.(2025)Ouyang, Yan, Hsu, Chen, Jiang, Wang, Han, Le,
  Daruki, Tang, Tirumalashetty, Lee, Rofouei, Lin, Han, Lee, and
  Pfister]{Ouyang2025ReasoningBankScalingAgentSelfEvolvingWith}
Siru Ouyang, Jun Yan, I-Hung Hsu, Yanfei Chen, Ke~Jiang, Zifeng Wang, Rujun
  Han, Long~T. Le, Samira Daruki, Xiangru Tang, Vishy Tirumalashetty, George
  Lee, Mahsan Rofouei, Hangfei Lin, Jiawei Han, Chen-Yu Lee, and Tomas Pfister.
\newblock Reasoningbank: Scaling agent self-evolving with reasoning memory.
\newblock \emph{CoRR}, abs/2509.25140, 2025.

\bibitem[Shenfeld et~al.(2026)Shenfeld, Damani, Hübotter, and
  Agrawal]{Shenfeld2026SelfDistillationEnablesContinualLearning}
Idan Shenfeld, Mehul Damani, Jonas Hübotter, and Pulkit Agrawal.
\newblock Self-distillation enables continual learning.
\newblock \emph{CoRR}, abs/2601.19897, 2026.

\bibitem[Shinn et~al.(2023)Shinn, Cassano, Gopinath, Narasimhan, and
  Yao]{Shinn2023ReflexionLanguageAgentsWithVerbalReinforcement}
Noah Shinn, Federico Cassano, Ashwin Gopinath, Karthik Narasimhan, and Shunyu
  Yao.
\newblock Reflexion: Language agents with verbal reinforcement learning.
\newblock In \emph{NeurIPS}, 2023.

\bibitem[Snell et~al.(2022)Snell, Klein, and
  Zhong]{Snell2022LearningByDistillingContext}
Charlie Snell, Dan Klein, and Ruiqi Zhong.
\newblock Learning by distilling context.
\newblock \emph{CoRR}, abs/2209.15189, 2022.

\bibitem[Wang et~al.(2025{\natexlab{a}})Wang, Yu, Gao, Zheng, Liu, Lu, Dang,
  Chen, Yang, Zhang, Liu, Yang, Zhao, Yue, Song, Yu, Huang, and
  Lin]{Wang2025Beyond8020Rule}
Shenzhi Wang, Le~Yu, Chang Gao, Chujie Zheng, Shixuan Liu, Rui Lu, Kai Dang,
  Xiong-Hui Chen, Jianxin Yang, Zhenru Zhang, Yuqiong Liu, An~Yang, Andrew
  Zhao, Yang Yue, Shiji Song, Bowen Yu, Gao Huang, and Junyang Lin.
\newblock Beyond the 80/20 rule: High-entropy minority tokens drive effective
  reinforcement learning for llm reasoning.
\newblock In \emph{Advances in Neural Information Processing Systems},
  volume~38, 2025{\natexlab{a}}.
\newblock \doi{10.52202/085713-3850}.
\newblock URL
  \url{https://proceedings.neurips.cc/paper_files/paper/2025/hash/a797c2d2e0c1fdabf4d1ab8cd0b465c6-Abstract-Conference.html}.

\bibitem[Wang et~al.(2026)Wang, Wang, Zeng, Zhang, Liu, Yang, Dai, Shi, Li,
  Tong, Hua, Zhang, and Zhang]{Wang2026FluxOPD}
Yuran Wang, Zekun Wang, Bohan Zeng, Ruixu Zhang, Wenxuan Liu, Liu Yang, Yifan
  Dai, Yang Shi, Bozhou Li, Chengzhuo Tong, Daili Hua, Yuanxing Zhang, and
  Wentao Zhang.
\newblock Flux-opd: On-policy distillation with evolving contexts.
\newblock \emph{CoRR}, abs/2607.28022, 2026.

\bibitem[Wang et~al.(2025{\natexlab{b}})Wang, Mao, Fried, and
  Neubig]{Wang2025AgentWorkflowMemory}
Zora~Zhiruo Wang, Jiayuan Mao, Daniel Fried, and Graham Neubig.
\newblock Agent workflow memory.
\newblock In \emph{ICML}, 2025{\natexlab{b}}.

\bibitem[White et~al.(2025)White, Dooley, Roberts, Pal, Feuer, Jain,
  Shwartz-Ziv, Jain, Saifullah, Dey, Shubh-Agrawal, Sandha, Naidu, Hegde,
  LeCun, Goldstein, Neiswanger, and Goldblum]{LiveBenchMath}
Colin White, Samuel Dooley, Manley Roberts, Arka Pal, Benjamin Feuer,
  Siddhartha Jain, Ravid Shwartz-Ziv, Neel Jain, Khalid Saifullah, Sreemanti
  Dey, Shubh-Agrawal, Sandeep~Singh Sandha, Siddartha~V. Naidu, Chinmay Hegde,
  Yann LeCun, Tom Goldstein, Willie Neiswanger, and Micah Goldblum.
\newblock Livebench: A challenging, contamination-limited llm benchmark.
\newblock In \emph{ICLR}, 2025.

\bibitem[Yang et~al.(2025)Yang, Li, Yang, Zhang, Hui, Zheng, Yu, Gao, Huang,
  Lv, Zheng, Liu, Zhou, Huang, Hu, Ge, Wei, Lin, Tang, Yang, Tu, Zhang, Yang,
  Yang, Zhou, Zhou, Lin, Dang, Bao, Yang, Yu, Deng, Li, Xue, Li, Zhang, Wang,
  Zhu, Men, Gao, Liu, Luo, Li, Tang, Yin, Ren, Wang, Zhang, Ren, Fan, Su,
  Zhang, Zhang, Wan, Liu, Wang, Cui, Zhang, Zhou, and
  Qiu]{Yang2025Qwen3TechnicalReport}
An~Yang, Anfeng Li, Baosong Yang, Beichen Zhang, Binyuan Hui, Bo~Zheng, Bowen
  Yu, Chang Gao, Chengen Huang, Chenxu Lv, Chujie Zheng, Dayiheng Liu, Fan
  Zhou, Fei Huang, Feng Hu, Hao Ge, Haoran Wei, Huan Lin, Jialong Tang, Jian
  Yang, Jianhong Tu, Jianwei Zhang, Jianxin Yang, Jiaxi Yang, Jing Zhou,
  Jingren Zhou, Junyang Lin, Kai Dang, Keqin Bao, Kexin Yang, Le~Yu, Lianghao
  Deng, Mei Li, Mingfeng Xue, Mingze Li, Pei Zhang, Peng Wang, Qin Zhu, Rui
  Men, Ruize Gao, Shixuan Liu, Shuang Luo, Tianhao Li, Tianyi Tang, Wenbiao
  Yin, Xingzhang Ren, Xinyu Wang, Xinyu Zhang, Xuancheng Ren, Yang Fan, Yang
  Su, Yichang Zhang, Yinger Zhang, Yu~Wan, Yuqiong Liu, Zekun Wang, Zeyu Cui,
  Zhenru Zhang, Zhipeng Zhou, and Zihan Qiu.
\newblock Qwen3 technical report.
\newblock \emph{CoRR}, abs/2505.09388, 2025.

\bibitem[Yang et~al.(2026)Yang, Wu, Lu, Shen, Zhang, Feng, Zhang, Luo, Lian,
  Wen, and Tao]{Yang2026OPIDOnPolicySkillDistillationFor}
Shuo Yang, Jinyang Wu, Zhengxi Lu, Yuhao Shen, Fan Zhang, Lang Feng, Shuai
  Zhang, Haoran Luo, Zheng Lian, Zhengqi Wen, and Jianhua Tao.
\newblock Opid: On-policy skill distillation for agentic reinforcement
  learning.
\newblock \emph{CoRR}, abs/2606.26790, 2026.

\bibitem[Yang et~al.(2018)Yang, Qi, Zhang, Bengio, Cohen, Salakhutdinov, and
  Manning]{HotPotQA}
Zhilin Yang, Peng Qi, Saizheng Zhang, Yoshua Bengio, William~W. Cohen, Ruslan
  Salakhutdinov, and Christopher~D. Manning.
\newblock Hotpotqa: A dataset for diverse, explainable multi-hop question
  answering.
\newblock In \emph{EMNLP}, pp.\  2369--2380, 2018.

\bibitem[Ye et~al.(2026{\natexlab{a}})Ye, Dong, Dong, Wu, Huang, and
  Wei]{Ye2026OnlineExperientialLearningForLanguageModels}
Tianzhu Ye, Li~Dong, Qingxiu Dong, Xun Wu, Shaohan Huang, and Furu Wei.
\newblock Online experiential learning for language models.
\newblock \emph{CoRR}, abs/2603.16856, 2026{\natexlab{a}}.

\bibitem[Ye et~al.(2026{\natexlab{b}})Ye, Dong, Wu, Huang, and
  Wei]{Ye2026OnPolicyContextDistillationForLanguage}
Tianzhu Ye, Li~Dong, Xun Wu, Shaohan Huang, and Furu Wei.
\newblock On-policy context distillation for language models.
\newblock \emph{CoRR}, abs/2602.12275, 2026{\natexlab{b}}.

\bibitem[Yuksekgonul et~al.(2025)Yuksekgonul, Bianchi, Boen, Liu, Lu, Huang,
  Guestrin, and Zou]{Yuksekgonul2025TextGrad}
Mert Yuksekgonul, Federico Bianchi, Joseph Boen, Sheng Liu, Pan Lu, Zhi Huang,
  Carlos Guestrin, and James Zou.
\newblock Optimizing generative ai by backpropagating language model feedback.
\newblock \emph{Nature}, 639\penalty0 (8055):\penalty0 609--616, 2025.

\bibitem[Zhang et~al.(2023)Zhang, Chen, Bukharin, Karampatziakis, He, Cheng,
  Chen, and Zhao]{Zhang2023AdaLoRA}
Qingru Zhang, Minshuo Chen, Alexander Bukharin, Nikos Karampatziakis, Pengcheng
  He, Yu~Cheng, Weizhu Chen, and Tuo Zhao.
\newblock {AdaLoRA}: Adaptive budget allocation for parameter-efficient
  fine-tuning.
\newblock In \emph{The Eleventh International Conference on Learning
  Representations}, 2023.
\newblock URL \url{https://openreview.net/forum?id=lq62uWRJjiY}.

\bibitem[Zhang et~al.(2025)Zhang, Hu, Upasani, Ma, Hong, Kamanuru, Rainton, Wu,
  Ji, Li, Thakker, Zou, and
  Olukotun]{Zhang2025AgenticContextEngineeringEvolvingContextsFor}
Qizheng Zhang, Changran Hu, Shubhangi Upasani, Boyuan Ma, Fenglu Hong,
  Vamsidhar Kamanuru, Jay Rainton, Chen Wu, Mengmeng Ji, Hanchen Li, Urmish
  Thakker, James Zou, and Kunle Olukotun.
\newblock Agentic context engineering: Evolving contexts for self-improving
  language models.
\newblock \emph{CoRR}, abs/2510.04618, 2025.

\bibitem[Zhao et~al.(2026)Zhao, Xie, Liu, Huang, Pang, Chen, and
  Grover]{Zhao2026SelfDistilledReasonerOnPolicySelf}
Siyan Zhao, Zhihui Xie, Mengchen Liu, Jing Huang, Guan Pang, Feiyu Chen, and
  Aditya Grover.
\newblock Self-distilled reasoner: On-policy self-distillation for large
  language models.
\newblock \emph{CoRR}, abs/2601.18734, 2026.

\end{thebibliography}
